\documentclass[10pt,journal,compsoc]{IEEEtran}
\ifCLASSOPTIONcompsoc
  \usepackage[nocompress]{cite}
\else
  \usepackage{cite}
\fi
\usepackage{times}
\usepackage{epsfig}
\usepackage{graphicx}
\usepackage{amsmath}
\usepackage{amssymb}
\usepackage{adjustbox}
\usepackage{booktabs}
\DeclareMathOperator*{\argmax}{arg\,max}

\DeclareMathAlphabet\mathbfcal{OMS}{cmsy}{b}{n}
\usepackage[colorinlistoftodos]{todonotes}

\ifCLASSINFOpdf
\else
\fi
\begin{document}
%
\title{Leveraging The Topological Consistencies of Learning in Deep Neural Networks}
%
%
%
%

\author{Stuart~Synakowski,Fabian Benitez-Quiroz,
        Aleix M. Martinez 
\IEEEcompsocitemizethanks{\IEEEcompsocthanksitem S. Synakowski F. Benitez-Quiroz, and A. Martinez are with the Department
of Electrical and Computer Engineering, The Ohio State University, Columbus,
OH, 43210.\protect\\
E-mail: synakowski.1@buckeyemail.osu.edu
}
\thanks{Manuscript received ---; revised ---}}

%
%

\markboth{October 2021 (Under Review)}%
{Shell \MakeLowercase{\textit{et al.}}: Bare Demo of IEEEtran.cls for Computer Society Journals}
%



\IEEEtitleabstractindextext{%
\begin{abstract}
 Recently, methods have been developed to accurately predict the testing performance of a Deep Neural Network (DNN) on a particular task, given statistics of its underlying topological structure. However, further leveraging this newly found insight for practical applications is intractable due to the high computational cost in terms of time and memory. In this work, we define a new class of \textit{topological features} that accurately characterize the progress of learning while being quick to compute during running time.
Additionally, our proposed topological features are readily equipped for backpropagation, meaning that they can be incorporated in end-to-end training.
Our newly developed \textit{practical} topological characterization of DNNs allows for an additional set of applications. We first show we can predict the performance of a DNN \textit{without a testing} set and without the need for high-performance computing.  We also demonstrate our topological characterization of DNNs is effective in estimating task-similarity. Lastly, we show we can induce learning in DNNs by actively constraining the DNN's topological structure. This opens up new avenues in constricting the underlying structure of DNNs in a meta-learning framework.
\end{abstract}

}


\maketitle

\IEEEdisplaynontitleabstractindextext

\IEEEpeerreviewmaketitle

%

\IEEEraisesectionheading{\section{Introduction}\label{sec:introduction}}
\IEEEPARstart{D}{eep}  neural networks (DNNs) are currently ubiquitous in the computer vision community due to their unprecedented performance in many computer vision tasks such as large-scale image classification problems \cite{deng2009imagenet}. However,
DNNs show success mainly in problems that require large amounts of labeled data for both training and performance evaluation. An open problem is to achieve the same level of performance using as little labeled data as possible. Paradigms like transfer-learning, few-shot-learning, and meta-learning all attempt to alleviate the problem of labeled training data. These paradigms are all similar in that they leverage a pre-trained embedding function or an additional objective function that relates to the current problem being solved. However, the aforementioned methods view DNNs more as an abstraction, with less consideration towards the underlying structure of the DNN itself. The current paradigms, unfortunately, perpetuate the convention that we view DNNs as ``black boxes", lacking the notion of explainability. This view is nevertheless understandable, given that the success of DNNs shatter conventional notions in statistical learning theory like the bias-variance trade-off \cite{belkin2019reconciling}. Since conventional statistical learning theory has not scaled to appropriately examine common deep learning models, recent works have asked whether more appropriate frameworks can be developed to understand the characteristics of learning in DNNs  \cite{sejnowski2020unreasonable}. 

\begin{figure}
\begin{center}
 \includegraphics[width=1.0\linewidth]{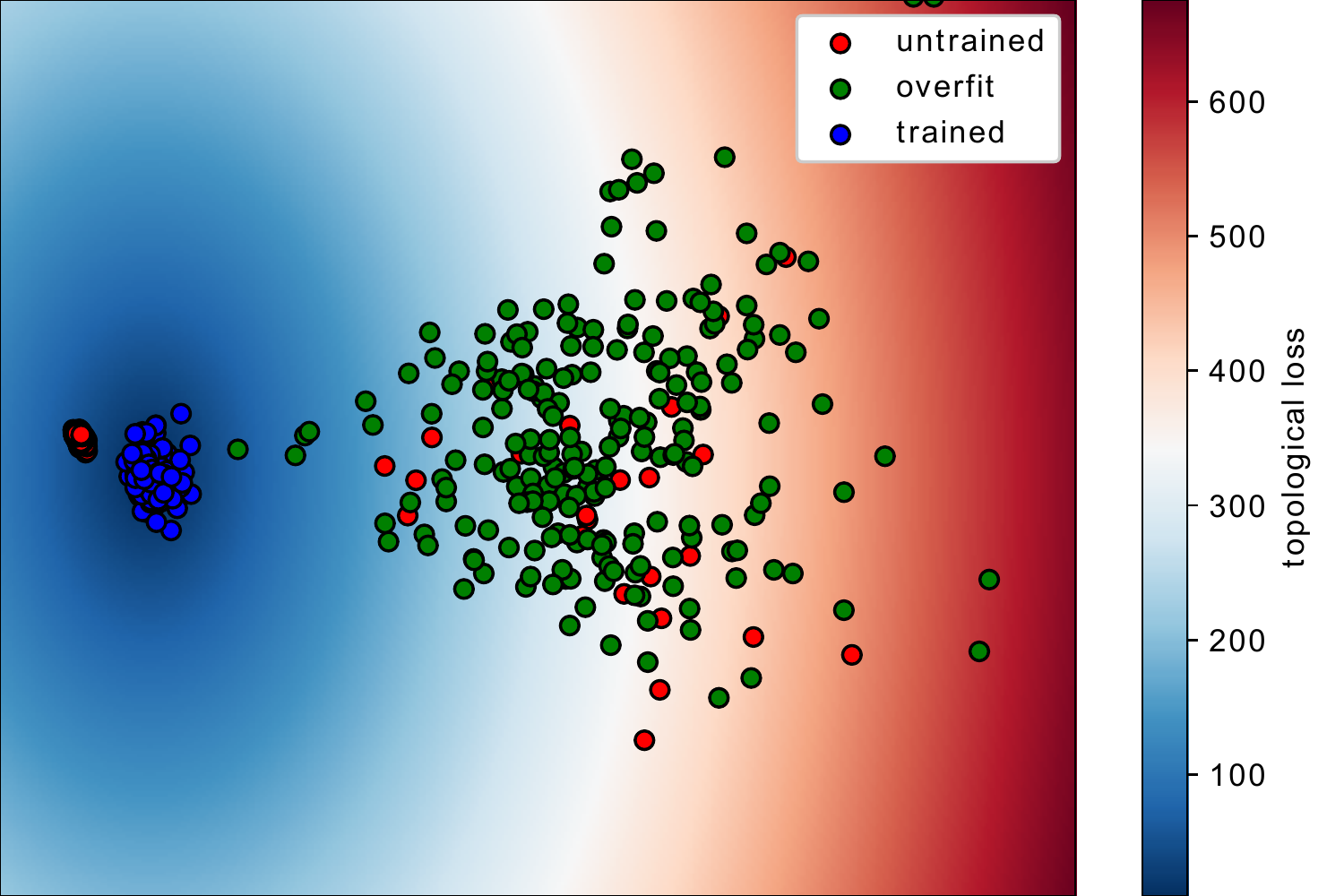}
\end{center}

  \caption{Shown above is a projection of what we refer to as the topological feature space of DNNs for a particular set of classification problems. Each point in this space denotes a network with its own set of parameters for a particular image classification task. This topological feature space effectively acts as a characterization of our hypothesis space. We see that regions in this space correspond to untrained, trained, and overfit models. In this work, we demonstrate the utility of such characterizations when they are computationally tractable. We not only show how we can achieve the goal of performance estimation without a testing set, but we can also infer the similarity between tasks and even construct meta-learning optimization strategies knowing the topological characteristics of learning.}

\label{fig:short}
\end{figure}



The difficulty in analyzing mathematical objects like DNNs is not only due to the high number of parameters, but the complex graphical structure needed to operate on high dimensional data. Fortunately, tools rooted in algebraic topology have been developed to examine and compare sophisticated high-dimensional mathematical objects like manifolds and graphs \cite{lum2013extracting}. The framework to examine persistent topological invariants is referred to as topological data analysis (TDA) and has shown to be an effective tool in understanding DNNs. Recent works have derived a set of features that effectively examine the underlying structure of DNNs that learn using TDA \cite{corneanu2019does,rieck2018neural,gabrielsson2019exposition}. 
Corneau et al. for example showed that trained DNNs elicit particular topological features which differ from untrained DNNs, allowing us to address issues regarding their interpretability \cite{corneanu2019does}.
The general framework of these methods is to construct a mapping $G$ from a deep neural network $\mathcal{F}$ and training dataset $\mathbf{X}$, to topological features $\mathbf{t}$. The topological feature representation $\mathbf{t}$ is embedded in a space that we will refer to as the \textit{topological feature space}.
Given many model parameterizations and datasets, this topological feature space can be explored and used for tangible applications. For example, topological features can be used to accurately estimate the gap between training and testing performance \cite{corneanu2020computing}. 

Though the topological characterization proposed in \cite{corneanu2020computing}  is effective in constructing a compressed representation that characterizes the progress of learning, it is difficult to use in practice due to computational complexity in terms of time and memory. Additionally, previously constructed topological characterizations are not readily equipped for backpropagation. This means that the DNN's topological structure could not be explicitly modified using standard optimization strategies.  

To solve the previous problems, we define a new class of topological features that characterize the progress of learning and can be easily computed at each training step. Additionally, our topological features are readily equipped for backpropagation. These two improvements result in major implications. We first demonstrate our topological characterization is still effectively estimating the performance of DNNs without the need for high-performance computing.  Additionally, we show how our understanding of the \textit{topological feature space} cannot only be used for performance estimation but for task-similarity estimation and meta-learning.

Our contributions are as follows
1) We define a new set of topological features, which characterize the topology of DNNs while being quick to compute at running time. This is especially useful when topological features of the network need to be analyzed at each iteration during training.   
2) We show that these topological features are effective at estimating the testing performance \emph{without the need of a testing set}.
3) We construct a mechanism to infer whether a model is appropriate for model selection given its representation in the topological feature space. 
4) We construct an optimization strategy that enforces the topological consistencies of learning. This is shown to improve testing performance, even when training on small datasets. 

\section{Related Work}
\label{sec:related}
The sub-fields of meta-learning task similarity and general performance improvement for deep learning have been heavily explored in the machine learning community. For this work, however,  we restrict ourselves to focusing on how topological data analysis has been applied in improving performance and understanding of DNNs.

\textbf{Understanding DNNs with TDA:} 
Topological data analysis (TDA) has demonstrated its use in understanding the topological differences between trained and untrained DNNs. \cite{corneanu2019does} observed homological differences between trained and untrained deep networks when viewing the network as a clique complex, where connections between nodes are defined by the correlation in activations between them. Additionally, recent works observed similar results when viewing weight matrices of each layer in a neural network as a filtration (sequence of simplicial complexes) and performing persistent homology on the weight matrices to construct a stopping criterion \cite{rieck2018neural}.  Other works observed that the convolutions in deep networks elicit consistent topological structures when viewing each filter as a point in a higher-dimensional space \cite{carlsson2018topological}.

These topological structures are so consistent that the testing error can be predicted as a function of them.  In \cite{corneanu2020computing}, the authors showed that the topological features induced by the correlation in activation between nodes can accurately predict testing error. Moreover, accurate predictions of the testing error were shown to be invariant across tasks, architectures, and datasets.
Topological features induced from deep networks have also been used to determine whether networks have been fed adversarial samples \cite{gebhart2019characterizing,corneanu2019does,gebhart2017adversary}.

\textbf{Comparing DNNs with TDA for Task Similarity}
Examining a DNNs topological structure as a function of its training task has only been explored very recently.
In \cite{tda_network_similarity}, the authors construct a topological feature representation as a function of weight matrices and compare how the topological similarity varies with respect to change of task and change of model architecture. This work does not, however, show if the topological feature representation can be used for predicting task similarity.

%
%



\textbf{Regularization and Model Selection with TDA:}
Recent works have developed methods that select model architectures as a function of the homology of the data itself. Intuitively, if the model cannot characterize the homology of the underlying dataset, the model certainly does not have the capacity to characterize the underlying structure of the data 
\cite{guss2018characterizing}.
Network pruning methods have been constructed as a function of topological features. In \cite{network_pruning} the authors first construct a filtration as a function of weight matrices. They then select weight values to be set to zero if those weight values correspond to certain homology classes within the filtration.
In addition, other works have selected operators within the deep networks as a function of their topological properties
\cite{bergomi2019towards}.
Preserving topological information in images has shown to be useful in deep image segmentation problems \cite{hu2019topology,haft2020topological,clough_TPAMIsegmentation_medical}.
Other works have applied differentiable frameworks which focus on the topology of the data being operated on. In \cite{moor2019topological} topological constraints in the latent space of an auto-encoder were introduced to obtain a more interpretable latent space. \cite{chen2018topological} applied topological constraints on the decision boundary of classifiers to mitigate overfitting.
However, these methods only consider topological characteristics of the feature representation and do not consider the consistent topological features elicited by the networks themselves.  Recently, frameworks have been developed to more conveniently induce specific topological features using gradient-based approaches \cite{bruel2019topology}. Instead of merely characterizing networks topologically, we can actually enforce topological changes within the network using the conventional optimization strategies in neural network training.

\section{Topological Preliminaries}
\label{sec:prelim}

Topological data analysis (TDA) characterizes the \textit{shape} of the data using tools derived from algebraic topology. In this work, data is presented as a set of points $S$ on a manifold $M$. Given a set of points $S \subset M \subset  \mathbb{R}^d $, we approximate the manifold by constructing a geometric simplicial complex $K_\psi(S)$, which is interpreted as a ``triangulation" of our manifold M induced by S and a distance threshold $\psi$, where  $\psi$ determines connections between points in S. 
A common topological feature extracted from $K_{\psi}(S) $ is its n-dimensional homology, and is often interpreted as characterizing the n-dimensional \textbf{\textit{voids}} in $K_{\psi}(S) $.  Since $K_{\psi}(S)$ changes with respect to the threshold $\psi$. It is of interest to see how the homology of $K_{\psi}(S) $ changes with respect to $\psi$. While varying the threshold $\psi$ we can construct a sequence of simplicial complexes known as a filtration $\mathbf{K}(S)$, where
\begin{equation}
\mathbf{K}(S) : = \emptyset \subseteq  K_{\psi_1} \subseteq  K_{\psi_2} \subseteq  K_{\psi_3}...
\end{equation}
Using the filtration $\mathbf{K}(S)$ we record the thresholds $\psi_i$ where the n-dimensional homology changes in the filtration. 
For analysis, we consider the thresholds $\psi_i$ where n-dimensional voids are created or filled, also known as the birth and death of homology classes. The birth and death of a homology class is denoted as a persistence point $(\psi_{birth},\psi_{death})$. It is the collection of persistent points that are used to characterize the shape of our data. We note that our filtration constructions from DNNs differ greatly from previous works which may view DNNs as non-euclidean data. See \cite{edelsbrunner2010computational} for rigorous definitions of abstract/ geometric simplicial complexes and persistent homology.
Moreover, there are many classes of filtered complexes induced by point-sets. In this work, we examine the persistent homology of weak-alpha filtrations described by Bruel-Gabrielsson et al. \cite{bruel2019topology}. We define $\mathbb{P}_n(S)$ as the set of persistent points corresponding to  n-dimensional homology classes within our weak-alpha filtration $\mathbf{K}(S)$. 

Many works have defined embeddings and similarity measures to analyze persistent points \cite{carriere2020perslay,pun2018persistent,bubenik2015statistical}.
For this work, we simply compute statistics on the death of homology classes of $\mathbb{P}_0(S)$  to determine a topological characterization $t_S$ of $S$. Hence,
\begin{equation}
t_S=g(\mathbb{P}_0(S))
\end{equation}
where $g$ is a differentiable function that computes statistics on persistence values. Such statistics include the maximum, minimum, mean, and standard deviation of $\mathbb{P}_n(S)$   
Thus, $g(\mathbb{P}_0(S))$ is the operation that performs topological data analysis on $S$.

When incorporating topological features into a learning framework, it is of interest to understand how to differentiate elements of $t_s$ with respect to points in $S$. We use the optimization strategy defined in \cite{bruel2019topology}. In \cite{bruel2019topology}, the derivative of topological features with respect to persistence values are equated to the derivative of topological features with respect to the pairs of points creating the persistence value. Hence given some topological statistic $t_{stat}$ and some threshold $\psi_i \in P_0(S)$, 
\begin{equation}
\frac{\partial t_{stat}}{\partial \psi_i}= \frac{\partial t_{stat}}{\partial d^{pair}_{\psi_i}}
\end{equation}
where $ d^{pair}_{\psi_i}$ denotes the distance between the pair of points in $S$ that created the death of a homology class. 
Intuitively, increasing the persistence of a homology class is induced by making local increases to the distances between the pair of points that destroyed it.

\begin{figure*}
\begin{center}
 \includegraphics[width=1.0\linewidth]{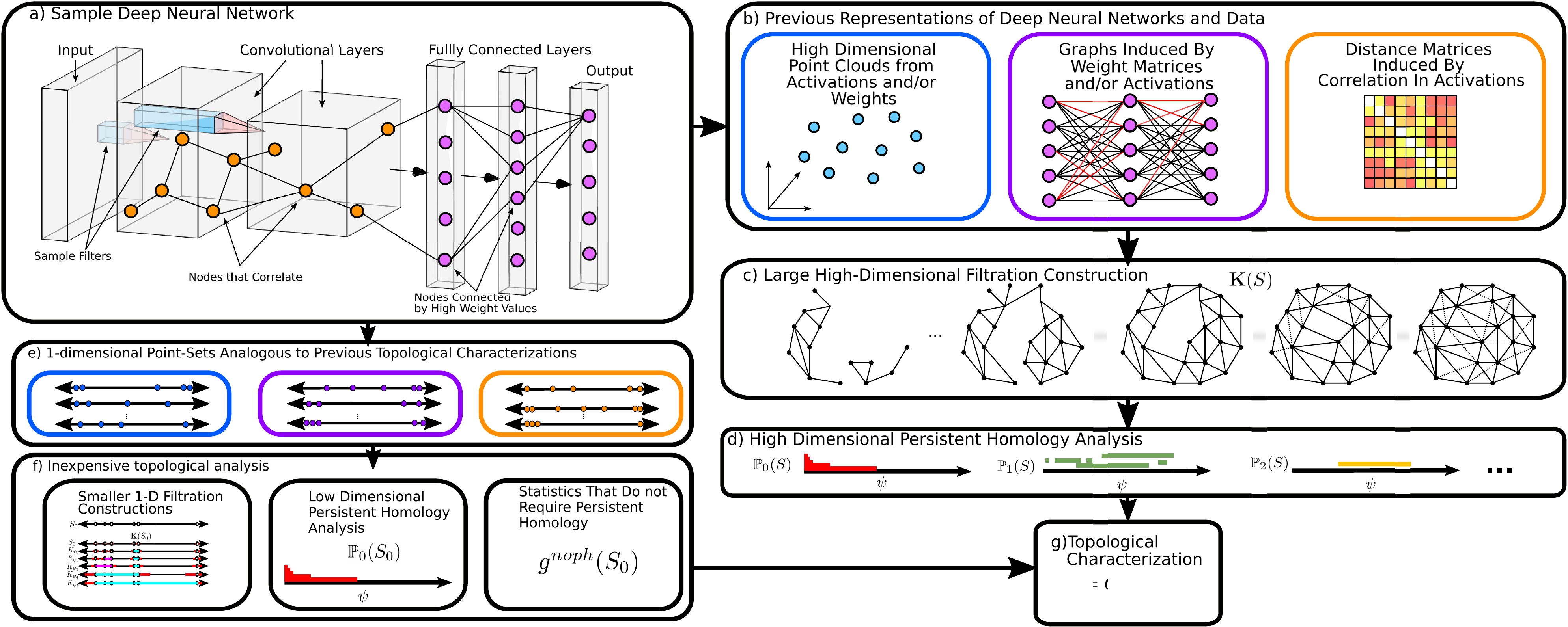}
\end{center}

  \caption{Shown above is a comparison between previous topological characterizations and ours, to provide some intuition for our filtration constructions. a) Given a particular deep neural network which in this case is a LeNet style architecture  b) Previous works represent deep neural networks by first defining high dimensional point clouds, large graphs, or large distance matrices, each induced by weight matrices or activation values.
  c-d) However the previous works construct filtrations that are computationally expensive to construct and analyze the persistent homology of to ultimately construct a topological characterization $\mathbf{t}_c$ shown in g). 
  e-f) In our work we construct a set of topological representations inspired by previous topological characterizations, but our topological characterizations are significantly less expensive computationally. This is because our topological characterizations operate on smaller 1-dimensional point-set representations of our deep neural network. Moreover, some of our operations do not require persistent homology analysis when producing $\mathbf{t}_c$.}

\label{fig:comaring_old}
\end{figure*}

\section{Methods}
Our first objective is to develop a topological characterization $\mathbf{t}_{\text{c}}$ of a  DNN $\mathcal{F}_\text{c}$ trained for a particular classification task $\mathcal{T}_{\text{c}}$, where $c$ denotes the task. Hence,
we aim to construct a differentiable mapping $G$ from our DNN and input data to our topological characterization of the network $\mathbf{t}_c$, i.e. 
\begin{equation}
\mathbf{t}_{\text{c}}=G(\mathcal{F}_{\text{c}},\mathbf{X}^{\text{train}}_{\text{c}}). 
\end{equation}
We then show how to use this embedding function to construct a performance estimation strategy, task-similarity estimation strategy, and a meta-learning/knowledge transfer strategy. 

\subsection{Network Topology}

We begin by defining the components of the network architecture, its parameters, and hidden feature representations of the DNN's input data. We denote a set of labeled training data for task $\mathcal{T}_{\text{c}}$ as $ D^{\text{train}}_{\text{c}}=\Big\{( X_{i},y_{i}) \Big \}_{i=1}^{|D^{\text{train}}_{\text{c}}|} $, where $X_i \in \mathbb{R}^{|X_{\text{c}}|}$ is the sample input vector of size $|X_{\text{c}}|$, and $y_i \in \mathbb{N}$ denotes the sample label. When training a deep network, training samples are represented as a tensor; thus, we define $ \mathbf{X}^{\text{ train}}_{\text{c}} \in \mathbb{R}^{n \times |X_{\text{c}}|} $ and $\mathbf{y}_{\text{c}}^{\text{train}} \in \mathbb{N}^{|D^{\text{train}}_{\text{c}}|}$ 
which concatenates all of the input training samples and training labels in $ X^{\text{train}}_{\text{c}}$ into single tensors. 

We denote our DNN $\mathcal{F}_{\text{c}}$ 
as a sequence of operators (layers) that iteratively transforms our input data $\mathbf{X}^{\text{train}}_{\text{c}}$ to find a representation that easily determines the class label. Thus,    
\begin{equation}
\mathcal{F}_{\text{c}} ( \mathbf{X}^{\text{train}}_{\text{c}}) =  f_L\circ  ... f_2 \circ f_{1} (\mathbf{X}^{\text{train}}_{\text{c}})
\end{equation}
or for a more convenient representation,
\begin{equation}
\mathcal{F}_{\text{c}} ( \mathbf{X}^{\text{train}}_{\text{c}}) := \odot_{i=1}^L f_i (\mathbf{h_i}^{\text{train}}_{\text{c}} )
\end{equation}where $\mathbf{h_{i}^{\text{train }}} \in \mathbb{R}^{|D^{\text{train}}_{\text{c}}| \times |h_l|} $,
denoting the hidden feature representation of n samples each of size $|h_i|$ fed into layer $f_i$.

For now, and without loss of generalization, we assume all layers are fully connected layers, and we examine convolutional layers in the supplementary material.  

We view the $i$-th fully connected layer of $\mathcal{F_{\text{c}}}$, $f_i$  as 
\begin{equation}
\mathbf{h_{i+1}^{\text{train}}} =f_i( \mathbf{h_{i}^{\text{train}}})=\phi_{\text{ReLU}}( \mathbf{h_{i}^{\text{train}}} W_i \oplus  b_i)
\end{equation}equipped with a weight matrix $W_l \in \mathbb{R}^{|h_i| \times |h_{i+1}|}$, bias term $b_i \in \mathbb{R}^{|h_{l+1}|} $ and an element-wise activation function 
$\phi_{\text{ReLU}}$ which is a  ReLU activation. $\oplus $ denotes a row-wise addition operator adding the bias term to the hidden feature representations of each sample. 
When considering a particular feature produced by layer $f_i$, we define $\mathbf{h_{i+1,j}^{\text{train }}} \in \mathbb{R}^{|D^{\text{train}}_{\text{c}}|}$ as the set of activations of the $j$-th column in $\mathbf{h_{i+1}^{\text{train }}} $ , also interpreted as the set of activations corresponding to the $j$-th node in the $i$+1-th layer of $\mathcal{F}_{\text{c}}$. 
We denote $w_{i,j,k}$ as the element in the $i$-th row and $j$-th column of $W_{i}$. We define the set of mean activations and standard deviation in activations of the $j$-th node in the $i$-th layer induced by $\mathbf{X}^{\text{train}}_{\text{c}}$ as $\mu^{\text{train}}_{i,j}$, $\sigma^{\text{train}}_{i,j}$ respectively.

\subsection{Topological Characterization}

\textbf{Filtrations Induced by 1-Dimensional point-sets.}
Previous works topologically characterize DNNs by examining the higher-dimensional persistent homology of filtrations induced by distance matrices. However, the run-time to compute the persistence points corresponding to $n$-dimensional homology classes grows exponentially with respect to $n$ \cite{corneanu2020computing}.
Moreover, constructing filtrations with the same number of points as nodes in an entire neural network is computationally expensive in both time and memory.  To address this issue we define our own filtrations inspired by previous works in \cite{corneanu2019does,corneanu2020computing,gebhart2019characterizing,rieck2018neural,gabella2019topology} which can be quickly computed. In addition, the features we compute are readily differentiable using methods described by Gabrielsson et al. \cite{bruel2019topology}.
Throughout our formulation of the framework, we will reference the previous works that inspired specific components of our topological characterization. Figure \ref{fig:comaring_old} provides some intuition relating previous topological characterizations and ours.

\textbf{General Filtration Construction:}
To alleviate the run-time and memory usage we consider the following conditions for our topological characterization:
1) We consider constructing filtrations that only require a low-dimensional persistent homology analysis.
2) We view the DNN as a set of smaller filtrations as opposed to characterizing the DNN as one large filtration.
3) We consider feature representations inspired by previous topological characterizations which do not require persistent homology analysis.

We accomplish the first condition by viewing the network and data as a set of weak-alpha filtrations induced by point-sets residing in  $\mathbb{R}$.  Given that these point-sets only reside in one dimension, we only examine $\psi$ thresholds corresponding to the death of 0-dimensional homology classes.  We then analyze the persistent homology of those weak-alpha filtrations and collect statistics on the persistence values to produce a topological characterization $\mathbf{t}_{\text{c}}$. Hence, given some point-set $S_0 \subset \mathbb{R}$, we extract a topological characterization using our operation $t_{S_0}=g(\mathbb{P}_0(S_0))$. For convenience, we denote the following operation as $g^{ph}:=g(\mathbb{P}_0)$. 

\begin{figure*}
\begin{center}
 \includegraphics[width=1.0\linewidth]{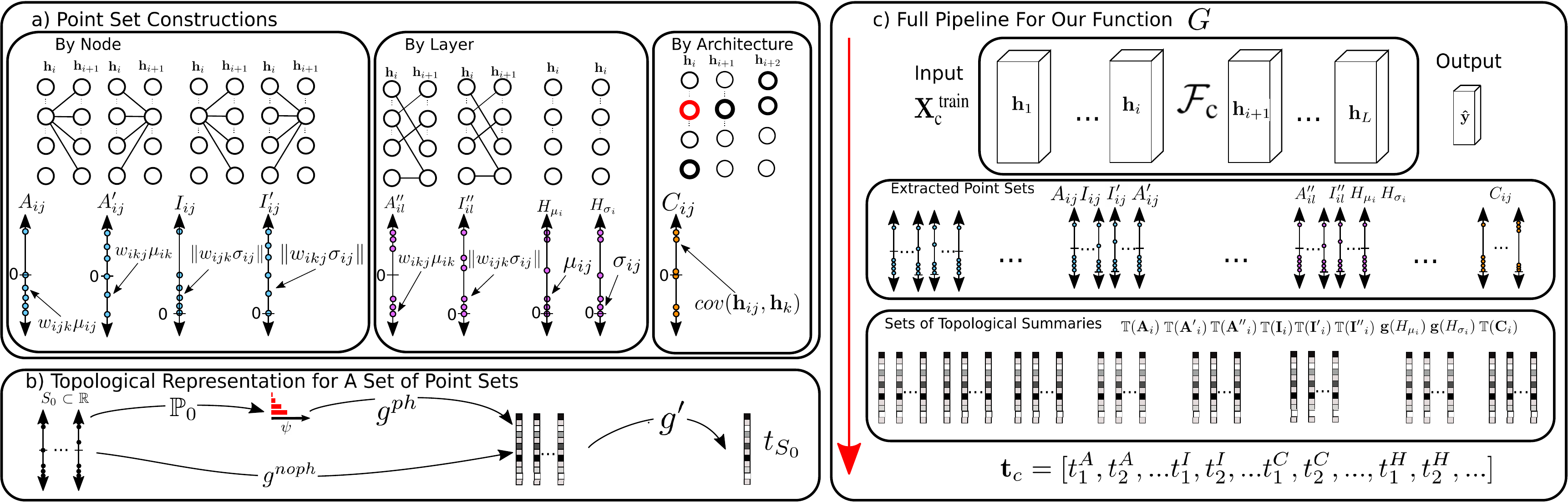}
   \caption{Schematic of how we topologically characterize the network. a) Here we show the classes of 1-dimensional point-sets extracted from the network, ranging from local to global structures inspired by previous works. b) From each of these point-sets we construct a topological characterization using our functions $g^{ph}$ and/or $g^{noph}$ along with $g'$.  c) We Show the full pipeline (top to bottom) to construct a topological characterization for a sample fully-connected network $\mathcal{F}_c$.}
 
\end{center}
\label{fig:forward_pass}
\end{figure*}

To accomplish the second condition we consider the following framework to ``break up" the DNN as a set of smaller filtrations.   
We propose a topological characterization that aims to capture the
local, intermediate, and global topological properties of the DNNs that learn, encompassing many related works that study DNNs with TDA. We study the topology of scaled weight connections from each node to the next, statistics of hidden activations for each layer, and the covariance between nodes across multiple layers. The schematic of our function $G$ is shown in Figure \ref{fig:forward_pass}. Depending on the size of our model $\mathcal{F}_c$, the classes of filtrations can be interchanged or removed to reduce the run time.

For the third condition, we examine statistics on the point-sets already provided by persistent homology analysis as this is already provided. Hence, we define $g^{noph}(S_0)$ as a set of statistics computed on the point-set $S_0$ itself rather than statistics of persistent homology classes induced by $S_0$. These features can then be concatenated to define $g^{both}(S_0)=[g^{ph}(S_0),g^{noph}(S_0)]$. 
 To be as general as possible with our future derivations, we define $\mathbf{g}$ to be a particular mapping chosen from $\{g^{ph},g^{noph},g^{both}\}$. We will see in our experiments how each class of topological characterization performs in various applications. See  figure \ref{fig:forward_pass} for a diagram of our point-set representations and overall pipeline in constructing our mapping $G$.

\textbf{The Topology of Local Structures in the DNN:}
The first class of topological characterizations are defined to examine the local structures in our model $\mathcal{F}_{c}$. We are interested in examining the behavior of nodes in the network as they operate on data from one layer to the next. Inspired by \cite{rieck2018neural}, we construct 1-dimensional sets of points by considering the topology of weight values connected to a particular node in a single layer.  We then scale these weight values by activation values like in \cite{gebhart2017adversary} and \cite{gebhart2019characterizing}. As opposed to scaling the weight values by a single activation value, we scale the weight values by statistics of the hidden activations provided by the training data, as this reduces the number of filtrations being constructed. We consider the mean and standard deviation of actions corresponding to a single node.
Consider the $j$-th node of the layer $f_i$ we define  
\begin{equation}
    A_{ij} = \{w_{ijk}\mu_{ij}|k\in 1...|h_{i+1}|\}
\end{equation}
and
\begin{equation}
    I_{ij} = \{\|w_{ijk}\sigma_{ij}\||k\in 1...|h_{i+1}|\}
\end{equation}
where $\|\cdot\|$ denotes the absolute value. 
These point-sets will be used to topologically characterize one node's contribution in manipulating the next hidden feature representation.
Similarly, we define
\begin{equation}
    A'_{ij} = \{w_{ikj}\mu_{ik}|k\in 1...|h_{i}|\}
\end{equation}
and
\begin{equation}
    I'_{ij} = \{\|w_{ikj}\sigma_{ij}\||k\in 1...|h_{i+1}|\}
\end{equation}
which are used to topologically characterize how the nodes in the previous layer impact the activations of a particular node in the next layer. We then analyze the point-sets using our previously defined topological characterization.
Hence, we define the following feature representations for the i-th layer,
\begin{equation}
\mathbb{T}(\mathbf{A}_i) = \{\mathbf{g}(A_{i1}),\mathbf{g}(A_{i2}) ...\}
\end{equation}
\begin{equation}
\mathbb{T}(\mathbf{A'}_i) = \{\mathbf{g}(A'_{i1}),\mathbf{g}(A'_{i2}) ...\}
\end{equation}
\begin{equation}
\mathbb{T}(\mathbf{I}_i) = \{\mathbf{g}(I_{i1}),\mathbf{g}(I_{l2}) ...\}
\end{equation}
\begin{equation}
\mathbb{T}(\mathbf{I'}_i )= \{\mathbf{g}(I'_{i1}),\mathbf{g}(I'_{i2}) ...\}
\end{equation}

\textbf{Topological Structures Induced by Layers:}
As previously discussed, recent works have constructed characterizations at higher resolutions by examining the topological structures induced by the entire layer.  To topologically characterize the entire layer we propose our own topological characterizations. We first randomly select weights from the weight matrix and scale them by their corresponding mean activations. Thus, for a single point-set,
\begin{equation}
\label{eq:16}
    A''_{il} = \{w_{ijk}\mu_{ij}| j\in J_l ,k \in K_l\}
\end{equation}
\begin{equation}
\label{eq:17}
    I''_{il} = \{||w_{ijk}\sigma_{ij}|| | i\in J_l ,j \in J_l\}
\end{equation}
Where $J_l$ and $K_l$ are a set of randomly selected indices.
We then compute the same topological feature representation as we did for the single node, hence
\begin{equation}
\mathbb{T}(\mathbf{A''}_{i}) = \{\mathbf{g}(A''_{i1}),\mathbf{g}(A''_{i2}) ...\}
\end{equation}
\begin{equation}
\mathbb{T}(\mathbf{I''}_{i}) = \{\mathbf{g}(I''_{i1}),\mathbf{g}(I''_{i2}) ...\}
\end{equation}

We also consider the topological characteristics of the hidden activations for each layer. Similar to \cite{gabella2019topology} we consider a lower-dimensional characterization of hidden activation values as data passes through the network. However, we simply consider statistics in the activation values.
\begin{equation}
H_{\sigma_i} = \{ \sigma_{ij} | j \in 1... |h_i|  \},
\end{equation}
and
\begin{equation}
H_{\mu_i} =  \{ \mu_{ij} | j \in ...|h_i|  \}
\end{equation} where the topological feature representation is
\begin{equation}
t^H_i = [\mathbf{g}(H_{\mu_i}),\mathbf{g}(H_{\sigma_i}) ].
\end{equation}

\textbf{Topological Structures Induced by the Entire Architecture}
To examine the global topological structure of the DNN, we consider the covariance in activations between nodes across multiple layers. Though a Rips-Filtration can be induced by distance matrices which are a function of covariance matrices like in \cite{corneanu2019does} and \cite{corneanu2020computing}, we drastically reduce the run time by considering the topology of point-sets induced by the covariance in activations between one node with respect to the activations of other nodes in the network. Given the set of activations for the j-th node in the i-th layer $\mathbf{h_{ij}}$, we define $C_{ij}$ as a set of covariance values between the $j$-th node in the $i$-th layer the activations from a set of nodes selected throughout the DNN. Thus
\begin{equation}
\label{eq:23}
C_{ij}=  \{cov(\mathbf{h}_{ij},\mathbf{h}_{k}) | k \in \mathcal{C}_{ij}^{\text{ind}}\},
\end{equation}
where $\mathcal{C}_{ij}^{\text{ind}}$ denotes a set of indices corresponding to nodes throughout the network. Thus for the i-th layer in the network 
\begin{equation}
\mathbb{T}(\mathbf{C}_i) = \{\mathbf{g}(C_{i1}),\mathbf{g}(C_{i2}) ...\}.
\end{equation}

\textbf{Aggregating Persistence Point Statistics into One Topological Feature}
Since we broke up our DNN $\mathcal{F}_c$ into a set of simplicial complexes at each layer, we need to ensure that our topological features are permutation invariant.
This is addressed by applying an additional set of statistics on our topological summaries. Similar to our function $g$, we denote $g'$ as a differentiable function which computes the mean and standard deviation of our sets of topological summaries. Thus for the $i$-th layer we have the following feature representations: 
\begin{equation}
t^{\text{A}}_{i}=[g'(\mathbb{T}(\mathbf{A}_i)),
g'(\mathbb{T}(\mathbf{A'}_i)),g'(\mathbb{T}(\mathbf{A''}_i))],
\end{equation}\begin{equation}
t^{\text{I}}_{i}=[g'(\mathbb{T}(\mathbf{I}_i)),
g'(\mathbb{T}(\mathbf{I'}_i)),g'(\mathbb{T}(\mathbf{I''}_i))],
\end{equation}
\begin{equation}
t^{\text{C}}_{i}=[g'(\mathbb{T}(\mathbf{C}_i))]
\end{equation}
We then combine all of our topological features into one vector for our final topological representation
\begin{equation}
\mathbf{t}_c = [t^A_{1},t^A_{2},... t^I_{1},t^I_{2},... t^C_{1},t^C_{2},...,t^H_{1},t^H_{2},... ].
\end{equation}

We define $G$ as the set of previously described operations that result in our topological representation $\mathbf{t}_c$. Since our topological feature representation is constructed using simplicial complexes equipped with gradients as described in \cite{bruel2019topology}, and the statistics we compute using g and g' are differentiable, we know that the composition of our differentiable functions $G$ is also differentiable. Therefore, we are equipped to enforce the network to elicit particular topological structures using conventional gradient-based optimization strategies.
\label{sec:methods}

\begin{figure*}
\begin{center}
 \includegraphics[width=1.0\linewidth]{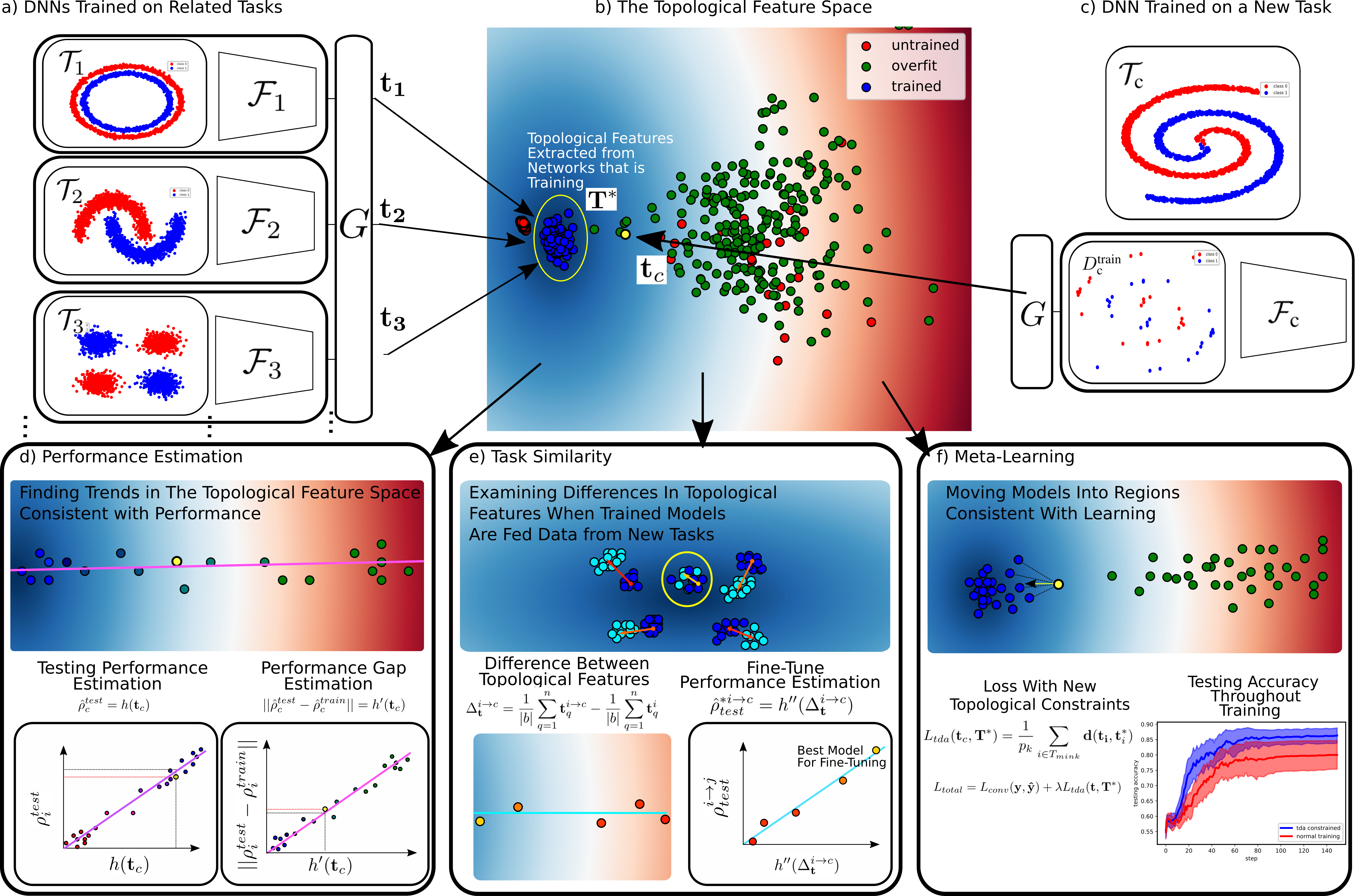}
\end{center}
  \caption{Here we show how we leverage our understanding of the topological feature space a) Shows a set of DNNs trained on a set of tasks where training data was abundant. b) From these trained deep networks and tasks, we extract topological features with our function mapping $G$ and use the topological features $\mathbf{T^*}$ induced from networks we know achieve a high classification accuracy. c) Shows a new classification task where data is not as abundant with a DNN currently being trained. d) We can first analyze the topological feature to estimate our testing performance for our model $\mathcal{F}_c$ using our training set only. e) We can also feed our previously trained models our training data from $\mathcal{T}_c$ and estimate the fine-tuning testing performance.  This allows us to infer task similarity by determining the pre-trained model with the highest estimated testing performance. f) Lastly, we can "pull" models into regions consistent with learning within the topological feature space and find that it can consistently improve testing performance for many different classification problems.
  }

\label{fig:exploration_and_app}
\end{figure*}

\section{Applications}
\label{sec:apps}
We first verify our new topological features are capable of performing tasks explored by previous works which study deep networks with TDA. One task, in particular, is the ability to discriminate between trained and overfit models. Next, we show that our topological features are effective in estimating the testing performance and performance gap (difference between training and testing performance) of models that have been trained. We then demonstrate how these topological features can be used for pre-trained model selection to achieve a higher testing performance after fine-tuning as this effectively determines if two tasks are related to each other.
Lastly, we develop a meta-learning strategy to enforce the topological characteristics consisting with learning, that is, we explicitly enforce networks to have similar topological structures to networks successfully trained on different tasks. All applications are shown in figure \ref{fig:exploration_and_app}.

\subsection{Performance Estimation}
For each performance estimation experiment, we have gathered a set of topological features induced by models trained across a set of tasks. For a given classification task, we compute topological features $\mathbf{t}_i$, their intended model state $m^{state}_i$ (whether the model is untrained, trained with sufficient data, or trained with insufficient data), and training and testing accuracies $\rho^{train}_i,\rho^{test}_i$ respectively.  Note that this does not include topological features induced from our a testing current task $\mathcal{T}_c$.  
\begin{equation}
D^{meta}_{top}=\{ (\mathbf{t}_i,m^{state}_i,\rho^{train}_i,\rho^{test}_i ) \}_{i=1}^{|D_{top}|}.
\label{eq:meta}
\end{equation}

\textbf{Distinguishing Between Model States}
We explore the space where our topological features reside to determine the regions within the topological space that correspond to networks that learn. This is done by predicting \textit{the intended model states} of topological features $\mathcal{T}_c$ using a k-nearest neighbor classifier $\mathbb{NN}_k$ trained with standardized samples $\mathbf{t}_i$ and labels  $m^{state}_i$ defined in \eqref{eq:meta}. Hence, 
\begin{equation}
\widehat{m}^{state}_c=\mathbb{NN}_k(\mathbf{t}_c).
\end{equation}
where $\widehat{m}^{state}_c$ denotes the predicted model state induced by the k-nearest neighbor classifier.



\textbf{Estimating Testing Performance}
To estimate the testing performance we construct a linear model 
\begin{equation}
\hat{\rho}^{test}_c = h(\mathbf{t}_c)= \mathbf{\beta}_h^T \mathbf{t}_c  + \beta_{h0} ,
\end{equation}
 where $\mathbf{\beta}_h \in \mathbb{R}^{|\mathbf{t}_c|}$ and $\beta_{h0}\in \mathbb{R}$ denote the linear model parameters, and $\hat{\rho}^{test}_c$ denotes the estimated testing accuracy. Our linear model fitted using  all topological features $\mathbf{t}_i$ and testing performancea $\rho^{test}_i $ from $D^{meta}_{top}$ where the training accuracy is above a threshold. We fit our model $h$ by minimizing the mean square error (MSE) with an L1-regularizer (LASSO Regression).

\textbf{Estimating Performance Gap}
Similarly, we construct another LASSO model to estimate the performance gap where 
\begin{equation}
||\hat{\rho}^{test}_c - \hat{\rho}^{train}_c|| = h'(\mathbf{t}_c) = \mathbf{\beta}_{h'}^T \mathbf{t}_c  + \beta_{h'0} ,
\end{equation}
\noindent where is $h'$ is a linear model fitted using topological features $\mathbf{t_i}$, and performance gap 
$ ||\rho^{test}_i - \rho^{train}_i|| $ from $D^{meta}_{top}$, in which the training accuracy is above a threshold. $\mathbf{\beta}_{h'}$ and $\beta_{h'0}$ denote the linear model parameters for $h'$.

\subsection{Task Similarity Estimation For Pre-trained Model Selection}
\label{sec:task_sim}
Upon further exploration of this topological feature space, we have found that pre-trained models differ topologically when they are fed training data corresponding to different tasks. This has raised the question as to whether we can use this ``topological disparity" to determine if a pre-trained model is appropriate for a given task.
 In this application we assume we have a bank of models $\mathbfcal{F}_s=\{\mathcal{F}_1,\mathcal{F}_2,\mathcal{F}_3 ...\}$ trained on on a set of tasks with corresponding training datasets $\mathbf{X}_s =\{\mathbf{X}_1,\mathbf{X}_2,\mathbf{X}_3...\}$.
Given our new task $\mathcal{T}_c$ with a set of training samples $\mathbf{X}_c$, we could train a new network $\mathcal{F}_c$ using our current training set $\mathbf{X}_c$; however, it may be of interest to simply select a pre-trained model from $\mathbfcal{F}_s$ that is trained on a task related to $\mathcal{T}_c$. Moreover, instead of immediately training a whole set of pre-trained models in $\mathbfcal{F}_s$ to determine which model is best suited for $\mathcal{T}_c$, we can first examine the topological features induced by each of the models as this is computationally less expensive and does not require a testing set for $\mathcal{T}_c$ .

 We can construct the following topological feature representations that feeds our current data into the previously defined models, hence 
$\mathbf{t}^{i \rightarrow c}=\mathbf{G}(\mathcal{F}_i,\mathbf{X}_c)$, where $\mathbf{t}^{i \rightarrow c}$ denotes the topological features induced by feeding model $\mathcal{F}_i$, our current dataset $\mathbf{X}_c$.  We formulate the problem of inferring task similarity selecting the most appropriate model $\mathcal{F}_i$ for fine-tuning. Hence we desire a model $H$ which performs the following operation
\begin{equation}
i^*= H({X_c,\mathbf{X}_s,\mathbfcal{F}_s)},
\end{equation}
Where $i^*$ denotes the index corresponding to the most appropriate model in $\mathbfcal{F}_s$ for task $\mathcal{T}_c$. 

One might ask how we determine the model that is most appropriate for fine-tuning. The proxy we use to infer task-simlarity is simply an estimation of the fine-tuning testing performance for a given pre-trained model. Similar to our previous models which estimate the testing performance and performance gap, we estimate the fine-tuning testing performance of a given model as a function of topological features. However, unlike the topological features derived in our previous models, we  examine the difference between topological feature representations. 

Suppose we have a set of $|b|$ topological features corresponding to $|b|$ different training batches fed into a model trained on task $\mathcal{T}_i$. We first compute the average difference between topological features induced by models fed data they were previously trained on and data corresponding to the current tasks. Hence 
\begin{equation}
\Delta_\mathbf{t}^{i\rightarrow c} = \frac{1}{|b|} \sum_{q=1}^n  \mathbf{t}^{i\rightarrow c}_q -  \frac{1}{|b|} \sum_{q=1}^n \mathbf{t}^i_q
\end{equation}We then use $\Delta_\mathbf{t}^{i\rightarrow c}$ to estimate the testing performance of $\mathcal{F}_i$ on task $\mathcal{T}_c$ after fine-tuning.

Like the previous performance estimation methods, we construct a lasso linear model as a function of our topological difference to estimate the average final testing performance of the fine-tuned model.
\begin{equation}
\hat{\rho}^{*i\rightarrow c}_{test} = h''(\Delta_\mathbf{t}^{i\rightarrow c}) =\beta^T_{h''} \Delta_\mathbf{t}^{i\rightarrow c} + \beta_{h0''} 
\end{equation}
Where $\hat{\rho}^{*i\rightarrow c}_{test}$ denotes the expected final testing performance after fine-tuning models on task $\mathcal{T}_c$ which were pre-trained on $\mathcal{T}_i$. Finally, our model selection is defined to be 
\begin{equation}
i^* = \argmax_i h''(\Delta_\mathbf{t}^{i\rightarrow c})
\end{equation}When training our model  $h''$ we use the the following dataset
\begin{equation}
D_{top}^{meta-sim}=\{(\rho^{i \rightarrow j}_{test},\Delta_\mathbf{t}^{i\rightarrow j})\}_{i,j \neq c } 
\end{equation}not including any data corresponding to task $\mathcal{T}_c$

\subsection{Meta-Learning Strategy}
\label{sec:meta}
During conventional training we define a loss function $L_{conv}(\mathcal{F}_{\text{c}}(\mathbf{X}^{\textbf{train}}_{\textbf{c}}),\mathbf{y}_{\text{c}}^{\text{train}}) 
$ that is a function of the DNN's output and labeled training data. We aim to improve the testing performance of $\mathcal{F}_{\text{c}}$ by ``pulling" our DNN into regions within the topological feature space that are consistent with learning (See Figure \ref{fig:exploration_and_app}). This is accomplished by minimizing the distance between our current topological features $\mathbf{t}_c$ and the topological features we know to be consistent with learning.  Suppose we select a set  of topological features $\mathbf{T^*}$ from $D^{meta}_{top}$ induced from other DNNs we know perform well on various tasks in terms of their testing accuracy.
\begin{equation}
\mathbf{T^*}=\{ \mathbf{t}^*_i \}_{i=1}^{|T*|}=\{ \mathbf{t}_i \in D^{meta}_{top} | \rho^{test}_i > \rho_{\text{thresh}} \} 
\end{equation}
We construct a topological loss by selecting the $p_k$ nearest topological features with respect to $\mathbf{t}_c$ from a random sample of $\mathbf{T^*}$ selected at each training set. Thus,
\begin{equation}
L_{tda}(\mathbf{t}_c,\mathbf{T^*}) = \frac{1}{p_k}\sum_{i \in T_{min k}}  \mathbf{d}(\mathbf{t_c},{\mathbf{t}}^*_i)
\end{equation}where $T_{mink}$ is the set of indices denoting the closest $p_k$ topological features $\mathbf{t}$ in a random sub-sample of $\mathbf{T}^*$, and $\mathbf{d}(\cdot,\cdot)$ denotes a weighted euclidean distance between $\mathbf{t_c}$ and $\mathbf{t_i^*}$. 
We determine weighting by first standardizing the features according to $D^{meta}_{top}$ and selecting components of the topological features that correlate with testing performance. More formally,
\begin{equation}
d(\mathbf{t_c},\mathbf{t}^*_i)=\frac{1}{|\mathbf{t}|} \sum_{j=1}^{|\mathbf{t}|} \frac{\mathbf{1}_j ||t_{cj}-t^*_{ij}||}{\sigma_j}
\end{equation}
where $t_{cj}$ and $t^*_{ij}$ denote the components of $\mathbf{t_c}$ and $\mathbf{t}^*_i$ respectively, $\sigma_j$ denotes the standard deviation of the $j$-th topological feature component in $D^{meta}_{top}$ , and
\begin{equation}
 \mathbf{1}_j = \begin{cases} 
      1 ,& ||corr(\mathbf{t}^j,\mathbf{p}^{test})|| \geq \tau_{corr}    \\
      0 ,& ||corr(\mathbf{t}^j,\mathbf{p}^{test})|| < \tau_{corr}
   \end{cases}
\end{equation}
where  $corr(\mathbf{t}^j,\mathbf{p}^{test})$ denote the pearson correlation between the $j$-th topological feature component in and the corresponding testing accuracy estimated in $D^{top}_{meta}$ and $\tau_{corr}$ is a threshold.
When training we consider the following loss function
\begin{equation}
L_{total} = L_{conv}(\mathbf{y}^{\text{train}}_c,\widehat{\mathbf{y}}^{\text{train}}_c)  + \lambda L_{tda}(\mathbf{t}_c,\mathbf{T^*}),
\end{equation}where $\lambda$ is an additional hyperparameter.

\section{Experiments}
\subsection{Dataset Construction}
We gather topological features across mutually exclusive classification problems.  We define 3 classes of parent datasets used in the experiments: synthetic 2-dimensional data, greyscale image data, and downsampled ImageNet data. 

\textbf{Synthetic 2-dimensional datasets:} Our synthetic 2-dimensional datasets consist of a set of binary classification problems generated from the scikit learn database \cite{scikit-learn}. Such datasets include discriminating between samples that resemble spirals, moons, circles, and Gaussian distributed data. For each classification problem, we generate 4000 training and testing samples for each binary classification problem. To have more diversity in our datasets, we apply a series of augmentations like rotations and non-isotropic scaling to each generated 2D dataset. We assume the augmentations result in the different classification problems for the performance estimation experiments. However, for the task-similarity and meta-learning experiments, we do not include augmented datasets when fitting $h''$ or selecting $\mathbf{T}^*$.

\textbf{greyscale Image datasets:} We also consider greyscale image  classification problems.  We combine MNIST \cite{lecun-mnisthandwrittendigit-2010},  KMNIST \cite{clanuwat2018deep} and EMNIST  \cite{cohen2017emnist} datasets. This aggregation would have a total number of 46 classes. We randomly partition each of the grey image datasets into groups of $m \in 2, 4, \text{and } 8$. This means, that we randomly select m classes from the augmented dataset to test the applications in Section \ref{sec:apps} and the rest of the classes are used for generating topological features for training our meta-learning applications. We state the specific class partitioning in the supplementary material. Note that when we split the data, we maintain the initial partitioning of training and testing samples for each class as defined by previous works.

 
\textbf{Downsampled Imagenet Data: } We also use a downsampled version of ImageNet known as Tiny-Imagenet \cite{wu2017tiny} consisting of 200 classes of colored images. Image classes are randomly partitioned into groups of 2 such that we have 100 2-class problems. Samples belonging to the original training and testing sets remain as defined by previous works. Both image samples are standardized by pixel intensity. Specific class partitions are defined in the supplementary material.

\subsection{Models}
For synthetic 2D classification problems, we trained networks with fully connected layers each equipped with 25 hidden units with ReLu activation functions. 
For the greyscale and downsampled ImageNet datasets, models are equipped with a LeNet style architecture. The models first begin with convolution layers, equipped ReLU activations, and with max pooling. After the convolution layers, two fully connected layers are applied. See Table \ref{table::one} which defines the list of architectures used for our performance estimation and meta-learning strategy.


\begin{table*}
\begin{center}
\begin{adjustbox}{max width=\textwidth}
\scalebox{1.0}{
\begin{tabular}{|l|c|}
\hline
Model Name  & Model Operation Sequence \\
\hline\hline

synth fc6  & $ \text{In} \rightarrow (\text{FC-25} \rightarrow \text{Relu} )_{\times 5} \rightarrow \text{FC-NC} \rightarrow \text{Out} $\\ \hline
synth fc8  & $ \text{In} \rightarrow (\text{FC-25} \rightarrow \text{Relu} )_{\times 7} \rightarrow \text{FC-NC} \rightarrow \text{Out} $\\ \hline
synth fc10  & $ \text{In} \rightarrow (\text{FC-25} \rightarrow \text{Relu} )_{\times 9} \rightarrow \text{FC-NC} \rightarrow \text{Out} $\\ \hline
gconv2  & $ \text{In} \rightarrow (\text{Conv3x3-8} \rightarrow \text{Relu} \rightarrow \text{MaxPool2x2})_{\times 2} \rightarrow  $\\
  & $  \text{FC-50} \rightarrow \text{Relu} \rightarrow \text{FC-NC} \rightarrow \text{Out} $\\
\hline
gconv3  & $ \text{In} \rightarrow (\text{Conv3x3-8} \rightarrow \text{Relu} \rightarrow \text{MaxPool2x2})_{\times 3} \rightarrow  $\\
  & $  \text{FC-50} \rightarrow \text{Relu} \rightarrow \text{FC-Nc} \rightarrow \text{Out} $\\
\hline
gconv4  & $ \text{In} \rightarrow (\text{Conv3x3-8} \rightarrow \text{Relu} \rightarrow \text{MaxPool2x2})_{\times 2} \rightarrow  (\text{Conv3x3-64} \rightarrow \text{Relu})_{\times 2} \rightarrow $\\
 & $  \text{FC-50} \rightarrow \text{Relu} \rightarrow \text{FC-NC}  \rightarrow \text{Out} $\\

\hline
tconv3  & $ \text{In} \rightarrow (\text{Conv3x3-8} \rightarrow \text{Relu} \rightarrow \text{MaxPool2x2})_{\times 3} \rightarrow  \text{Conv3x3-16} \rightarrow \text{Relu} \rightarrow \text{MaxPool2x2} \rightarrow   $\\
 & $   \text{FC-50} \rightarrow \text{Relu} \rightarrow \text{FC-NC} \rightarrow \text{Out} $\\

\hline
tconv4  & $ \text{In} \rightarrow (\text{Conv3x3-8} \rightarrow \text{Relu} \rightarrow \text{MaxPool2x2})_{\times 2} \rightarrow \text{Conv3x3-16} \rightarrow \text{Relu} \rightarrow \text{MaxPool2x2} \rightarrow  \text{Conv3x3-16} \rightarrow \text{Relu} \rightarrow $\\ 
 & $   \text{FC-120} \rightarrow \text{Relu} \rightarrow \text{FC-NC} \rightarrow \text{Out} $\\

\hline
tconv5  & $ \text{In} \rightarrow (\text{Conv3x3-8} \rightarrow \text{Relu} \rightarrow \text{MaxPool2x2})_{\times 2} \rightarrow \text{Conv3x3-16} \rightarrow \text{Relu} \rightarrow \text{MaxPool2x2}  $\\
&$\rightarrow(\text{Conv3x3-16} \rightarrow \text{Relu} )_{\times 2}\rightarrow$  \\
& $ \text{FC-120} \rightarrow \text{Relu} \rightarrow \text{FC-NC} \rightarrow \text{Out} $\\

\hline
\end{tabular}}
 \end{adjustbox}
\end{center}
\caption{List of architectures used in our experiments. Models are denoted as a sequence of operations from left to right where each operation is denoted between arrows $\rightarrow$. FC and Conv, MaxPool, and Relu  denote fully-connected Layers, 2D-convolutional layers, max pooling layers and ReLu activation functions respectively. Each convolutional layer denotes the filter patch size as well as the number of filters used in each layer. For example, Conv3x3-8 denotes a convolutional layer equipped with 8 filters each filter being of 3x3. Similarly, MaxPool2x2 denotes a Max Pooling Layer with a kernel size of 2x2. FC-\# denotes a fully connected layer whose output representation is of size \# (NC denotes the number of classes for the particular classification problem).In and Out denote the input and output tensors respectively.$ _\# $ denotes a repetition of a sequence of operations. For example $(\text{FC-25} \rightarrow \text{Relu} )_{\times 5}$ would consist of a sequence of 5 fully-connected layers where each layer is equipped with a ReLU activation function.   }
\label{table::one}
\end{table*}

\begin{table*}[p] 
\centering 
\scalebox{0.90}{
\begin{tabular}{l | c c c | c c c | c c c }
  &  \multicolumn{9}{c}{\textbf{Testing Performance Estimation}} 
\\  \cmidrule(l){1-10} \textbf{Synthetic}  &  \multicolumn{3}{c}{\textbf{Model State}} & \multicolumn{3}{c}{\textbf{Testing Acc}} & \multicolumn{3}{c}{\textbf{Performance Gap}}
\\ \textbf{2D-Data}   &  \multicolumn{3}{c}{Mean Acc (\%)} & \multicolumn{3}{c}{ MAE (\%)} & \multicolumn{3}{c}{MAE (\%)}

\\ \textbf{Model}  & $g^{ph}$ & $g^{noph}$ & $g^{both}$ & $g^{ph}$ &  $g^{noph}$ & $g^{both}$ & $g^{ph}$ & $g^{noph}$ & $g^{both}$ 

\\ \cmidrule(l){1-1} \cmidrule(l){2-10} 
synth fc6 & 84.67 $\pm$ 2.79 &  88.33 $\pm$ 2.49 &  88.44 $\pm$ 2.67 &  5.04 $\pm$ 2.79 &  5.11 $\pm$ 2.49 &  4.97 $\pm$ 2.67 &  5.16 $\pm$ 0.65 &  5.11 $\pm$ 0.63 &  5.08 $\pm$ 0.62
\\ synth fc8 & 79.78 $\pm$ 3.06 &  83.00 $\pm$ 2.72 &  82.56 $\pm$ 2.94 &  5.10 $\pm$ 3.06 &  5.11 $\pm$ 2.72 &  4.90 $\pm$ 2.94 &  5.24 $\pm$ 0.62 &  5.15 $\pm$ 0.59 &  4.99 $\pm$ 0.59
\\   synth fc10 & 82.89 $\pm$ 2.65 &  87.00 $\pm$ 1.85 &  87.33 $\pm$ 2.00 &  4.98 $\pm$ 2.65 &  4.89 $\pm$ 1.85 &  4.88 $\pm$ 2.00 &  4.98 $\pm$ 0.51 &  4.86 $\pm$ 0.51 &  4.87 $\pm$ 0.51

\\  \cmidrule(l){1-1} \cmidrule(l){2-10} \textbf{Grey 2}  &  \multicolumn{3}{c}{\textbf{Model State}} & \multicolumn{3}{c}{\textbf{Testing Acc }} & \multicolumn{3}{c}{\textbf{Performance Gap}}
\\ \textbf{Data}  &  \multicolumn{3}{c}{Mean Acc (\%)} & \multicolumn{3}{c}{ MAE (\%)} & \multicolumn{3}{c}{MAE (\%)}
\\ \textbf{Model}  & $g^{ph}$ & $g^{noph}$ & $g^{both}$ & $g^{ph}$ &  $g^{noph}$ & $g^{both}$ & $g^{ph}$ & $g^{noph}$ & $g^{both}$ 
\\ \cmidrule(l){1-1} \cmidrule(l){2-10} 
gconv2 &  99.86 $\pm$ 0.14 &  99.86 $\pm$ 0.14 &  99.86 $\pm$ 0.14 &  3.65 $\pm$ 0.14 
&  3.61 $\pm$ 0.14 &  3.62 $\pm$ 0.14 &  3.49 $\pm$ 0.28 &  3.45 $\pm$ 0.27 &  3.45 $\pm$ 0.27

\\ gconv3 &  99.71 $\pm$ 0.20 &  99.71 $\pm$ 0.20 &  99.71 $\pm$ 0.20 &  3.60 $\pm$ 0.20 
&  3.53 $\pm$ 0.20 &  3.54 $\pm$ 0.20 &  3.60 $\pm$ 0.37 &  3.49 $\pm$ 0.35 &  3.50 $\pm$ 0.35

\\ gconv4 &  99.57 $\pm$ 0.23 &  99.71 $\pm$ 0.20 &  99.71 $\pm$ 0.20 &  4.07 $\pm$ 0.23 
&  3.84 $\pm$ 0.20 &  3.85 $\pm$ 0.20 &  3.77 $\pm$ 0.33 &  3.49 $\pm$ 0.31 &  3.49 $\pm$ 0.31

 \\  \cmidrule(l){1-1} \cmidrule(l){2-10} \textbf{Grey 4}  &  \multicolumn{3}{c}{\textbf{Model State}} & \multicolumn{3}{c}{\textbf{Testing Acc }} & \multicolumn{3}{c}{\textbf{Performance Gap}}
\\ \textbf{Data}  &  \multicolumn{3}{c}{Mean Acc (\%)} & \multicolumn{3}{c}{ MAE (\%)} & \multicolumn{3}{c}{MAE (\%)}
\\ \textbf{Model}  & $g^{ph}$ & $g^{noph}$ & $g^{both}$ & $g^{ph}$ &  $g^{noph}$ & $g^{both}$ & $g^{ph}$ & $g^{noph}$ & $g^{both}$ 

\\ \cmidrule(l){1-1} \cmidrule(l){2-10} 

gconv2 &  100.00 $\pm$ 0.00 &  100.00 $\pm$ 0.00 &  100.00 $\pm$ 0.00 &  4.98 $\pm$ 0.
00 &  5.04 $\pm$ 0.00 &  5.06 $\pm$ 0.00 &  4.19 $\pm$ 0.63 &  3.37 $\pm$ 0.48 &  3.37 $\pm$ 0.48
 
\\ gconv3 &  100.00 $\pm$ 0.00 &  99.70 $\pm$ 0.29 &  100.00 $\pm$ 0.00 &  4.19 $\pm$ 0.0
0 &  3.72 $\pm$ 0.29 &  3.69 $\pm$ 0.00 &  4.28 $\pm$ 0.65 &  3.56 $\pm$ 0.33 &  3.52 $\pm$ 0.34

\\ gconv4 &  100.00 $\pm$ 0.00 &  100.00 $\pm$ 0.00 &  100.00 $\pm$ 0.00 &  4.35 $\pm$ 0.
00 &  4.23 $\pm$ 0.00 &  4.29 $\pm$ 0.00 &  4.11 $\pm$ 0.58 &  3.73 $\pm$ 0.50 &  3.77 $\pm$ 0.50

 \\  \cmidrule(l){1-1} \cmidrule(l){2-10} \textbf{Grey 8}  &  \multicolumn{3}{c}{\textbf{Model State}} & \multicolumn{3}{c}{\textbf{Testing Acc }} & \multicolumn{3}{c}{\textbf{Performance Gap}}
\\ \textbf{Data}  &  \multicolumn{3}{c}{Mean Acc (\%)} & \multicolumn{3}{c}{ MAE (\%)} & \multicolumn{3}{c}{MAE (\%)}
\\ \textbf{Model}  & $g^{ph}$ & $g^{noph}$ & $g^{both}$ & $g^{ph}$ &  $g^{noph}$ & $g^{both}$ & $g^{ph}$ & $g^{noph}$ & $g^{both}$ 
\\ \cmidrule(l){1-1} \cmidrule(l){2-10}

gconv2 &  100.00 $\pm$ 0.00 &  100.00 $\pm$ 0.00 &  100.00 $\pm$ 0.00 &  4.60 $\pm$ 0.
00 &  3.77 $\pm$ 0.00 &  3.85 $\pm$ 0.00 &  4.55 $\pm$ 1.14 &  3.90 $\pm$ 0.99 &  3.94 $\pm$ 0.9

\\ gconv3 &  100.00 $\pm$ 0.00 &  99.33 $\pm$ 0.60 &  100.00 $\pm$ 0.00 &  3.59 $\pm$ 0.0
0 &  3.66 $\pm$ 0.60 &  3.68 $\pm$ 0.00 &  3.85 $\pm$ 0.96 &  4.05 $\pm$ 0.78 &  4.01 $\pm$ 0.78

\\ gconv4 &  99.33 $\pm$ 0.60 &  100.00 $\pm$ 0.00 &  99.33 $\pm$ 0.60 &  4.70 $\pm$ 0.60
 &  3.44 $\pm$ 0.00 &  3.66 $\pm$ 0.60 &  4.63 $\pm$ 0.86 &  3.41 $\pm$ 0.66 &  3.57 $\pm$ 0.66
 
 \\  \cmidrule(l){1-1} \cmidrule(l){2-10} \textbf{Tiny 2}  &  \multicolumn{3}{c}{\textbf{Model State}} & \multicolumn{3}{c}{\textbf{Testing Acc }} & \multicolumn{3}{c}{\textbf{Performance Gap}}
\\ \textbf{Data}  &  \multicolumn{3}{c}{Mean Acc (\%)} & \multicolumn{3}{c}{ MAE (\%)} & \multicolumn{3}{c}{MAE (\%)}
\\ \textbf{Model}  & $g^{ph}$ & $g^{noph}$ & $g^{both}$ & $g^{ph}$ &  $g^{noph}$ & $g^{both}$ & $g^{ph}$ & $g^{noph}$ & $g^{both}$ 
\\ \cmidrule(l){1-1} \cmidrule(l){2-10} 
tconv3 &  97.67 $\pm$ 0.35 &  98.80 $\pm$ 0.23 &  98.43 $\pm$ 0.26 &  6.66 $\pm$ 0.35 
&  6.57 $\pm$ 0.23 &  6.60 $\pm$ 0.26 &  6.01 $\pm$ 0.30 &  5.76 $\pm$ 0.29 &  5.76 $\pm$ 0.29

\\ tconv4 &  95.90 $\pm$ 0.51 &  97.87 $\pm$ 0.34 &  97.20 $\pm$ 0.43 &  7.26 $\pm$ 0.51 
&  7.12 $\pm$ 0.34 &  7.10 $\pm$ 0.43 &  6.88 $\pm$ 0.38 &  6.49 $\pm$ 0.36 &  6.49 $\pm$ 0.36

\\ tconv5 &  95.90 $\pm$ 0.45 &  97.53 $\pm$ 0.33 &  97.30 $\pm$ 0.33 &  7.42 $\pm$ 0.45 
&  7.09 $\pm$ 0.33 &  7.02 $\pm$ 0.33 &  7.06 $\pm$ 0.39 &  6.62 $\pm$ 0.36 &  6.59 $\pm$ 0.36

\\  \midrule
\end{tabular}}
\caption{Performance Estimation Results. The first 3 columns denote the average classification accuracy ( $\pm$ standard error) in predicting the intended model state using $\mathbb{NN}_k$ . The next 6 columns present the mean absolute error in \% $\pm$ the standard error in estimating the testing accuracy and performance gap respectively using $h$ and $h'$. $g^{ph}$,$g^{noph}$, and $g^{both}$ refer the particular topological characterization selected. } 
\label{tab:performance_tab}
\label{tab:three}

\end{table*}

\begin{table*}[h] 
\centering 
\scalebox{0.90}{
\begin{tabular}{l | c c c | c c c | c c c }
  &  \multicolumn{9}{c}{\textbf{Task Similarity}} 
\\  \cmidrule(l){1-10} \textbf{Synthetic}  &  \multicolumn{3}{c}{\textbf{Model Rank}} & \multicolumn{3}{c}{\textbf{Corr}} & \multicolumn{3}{c}{\textbf{Testing Improvement}}  
\\ \textbf{2D-Data}   &  \multicolumn{3}{c}{--} &  \multicolumn{3}{c}{--}  & \multicolumn{3}{c}{ (\%)} 

\\ $n_{models}=5$  & $g^{ph}$ & $g^{noph}$ & $g^{both}$ & $g^{ph}$ &  $g^{noph}$ & $g^{both}$ & $g^{ph}$ & $g^{noph}$ & $g^{both}$ 
\\ \cmidrule(l){1-1} \cmidrule(l){2-10} 
synth fc6 & 2.00 $\pm$ 0.28 &  1.60 $\pm$ 0.36 &  1.60 $\pm$ 0.36 &  0.41 $\pm$ 0.26 &  0.52 $\pm$ 0.26 &  0.51 $\pm$ 0.25 &  2.59 $\pm$ 1.66 &  4.07 $\pm$ 1.77 &  4.07 $\pm$ 1.77
\\ synth fc8 & 1.80 $\pm$ 0.44 &  1.40 $\pm$ 0.36 &  1.80 $\pm$ 0.44 &  0.46 $\pm$ 0.24 &  0.69 $\pm$ 0.15 &  0.68 $\pm$ 0.15 &  3.68 $\pm$ 1.79 &  4.21 $\pm$ 1.67 &  3.68 $\pm$ 1.79
\\ synth fc10 & 2.20 $\pm$ 0.33 &  1.20 $\pm$ 0.18 &  1.20 $\pm$ 0.18 &  0.65 $\pm$ 0.08 &  0.87 $\pm$ 0.06 &  0.88 $\pm$ 0.03 &  2.96 $\pm$ 2.41 &  5.64 $\pm$ 1.68 &  5.64 $\pm$ 1.68

\\  \cmidrule(l){1-1} \cmidrule(l){2-10} \textbf{Grey 2}  &  \multicolumn{3}{c}{\textbf{Model Rank}} &  \multicolumn{3}{c}{\textbf{Corr}}  &  \multicolumn{3}{c}{\textbf{Testing Improvement}}
\\ \textbf{2D-Data}   &  \multicolumn{3}{c}{--} & \multicolumn{3}{c}{--} & \multicolumn{3}{c}{(\%)}
\\ $n_{models}=22$  & $g^{ph}$ & $g^{noph}$ & $g^{both}$ & $g^{ph}$ &  $g^{noph}$ & $g^{both}$ & $g^{ph}$ & $g^{noph}$ & $g^{both}$ 
\\ \cmidrule(l){1-1} \cmidrule(l){2-10} 
 gconv2 &  7.22 $\pm$ 1.04 &  7.61 $\pm$ 1.14 &  7.39 $\pm$ 1.13 &  0.27 $\pm$ 0.05 &  0.39 $\pm$ 0.04 &  0.40 $\pm$ 0.04 &  2.68 $\pm$ 0.63 &  2.68 $\pm$ 0.75 &  2.78 $\pm$ 0.74
 \\ gconv3 &  8.39 $\pm$ 1.12 &  7.13 $\pm$ 0.85 &  7.30 $\pm$ 0.92 &  0.38 $\pm$ 0.04 &  0.41 $\pm$ 0.04 &  0.46 $\pm$ 0.04 &  2.07 $\pm$ 0.76 &  3.23 $\pm$ 0.67 &  3.10 $\pm$ 0.70
 \\ gconv4 &  8.13 $\pm$ 0.94 &  6.04 $\pm$ 1.06 &  6.13 $\pm$ 1.05 &  0.63 $\pm$ 0.04 &  0.66 $\pm$ 0.04 &  0.70 $\pm$ 0.03 &  4.22 $\pm$ 0.84 &  5.69 $\pm$ 1.03 &  5.65 $\pm$ 0.99
 
 \\  \cmidrule(l){1-1} \cmidrule(l){2-10} \textbf{Grey 4}  &   \multicolumn{3}{c}{\textbf{Model Rank}} & \multicolumn{3}{c}{\textbf{Corr}} & \multicolumn{3}{c}{\textbf{Testing Improvement}} 
\\ \textbf{Data}  &  \multicolumn{3}{c}{--} & \multicolumn{3}{c}{--} & \multicolumn{3}{c}{(\%)}
\\ $n_{models}=11$   & $g^{ph}$ & $g^{noph}$ & $g^{both}$ & $g^{ph}$ &  $g^{noph}$ & $g^{both}$ & $g^{ph}$ & $g^{noph}$ & $g^{both}$ 

\\ \cmidrule(l){1-1} \cmidrule(l){2-10} 
 gconv2 &  3.36 $\pm$ 0.78 &  2.00 $\pm$ 0.39 &  2.00 $\pm$ 0.39 &  0.53 $\pm$ 0.10 &  0.71 $\pm$ 0.06 &  0.76 $\pm$ 0.06 &  3.70 $\pm$ 1.07 &  5.38 $\pm$ 0.74 &  5.49 $\pm$ 0.81
 \\ gconv3 &  4.55 $\pm$ 0.80 &  2.27 $\pm$ 0.37 &  2.09 $\pm$ 0.35 &  0.55 $\pm$ 0.10 &  0.69 $\pm$ 0.07 &  0.75 $\pm$ 0.06 &  1.70 $\pm$ 1.02 &  5.08 $\pm$ 0.73 &  5.70 $\pm$ 0.71
 \\ gconv4 &  3.91 $\pm$ 0.66 &  3.27 $\pm$ 0.41 &  3.45 $\pm$ 0.61 &  0.58 $\pm$ 0.06 &  0.74 $\pm$ 0.05 &  0.79 $\pm$ 0.04 &  3.30 $\pm$ 1.34 &  4.85 $\pm$ 1.05 &  4.67 $\pm$ 1.14

 \\  \cmidrule(l){1-1} \cmidrule(l){2-10} \textbf{Grey 8}  &   \multicolumn{3}{c}{\textbf{Model Rank}} & \multicolumn{3}{c}{\textbf{Corr}} & \multicolumn{3}{c}{\textbf{Testing Improvement}}  
\\ \textbf{Data}  &  \multicolumn{3}{c}{--} & \multicolumn{3}{c}{--} & \multicolumn{3}{c}{(\%)}
\\ $n_{models}=5$  & $g^{ph}$ & $g^{noph}$ & $g^{both}$ & $g^{ph}$ &  $g^{noph}$ & $g^{both}$ & $g^{ph}$ & $g^{noph}$ & $g^{both}$ 
\\ \cmidrule(l){1-1} \cmidrule(l){2-10} 
gconv2 &  1.20 $\pm$ 0.18 &  1.00 $\pm$ 0.00 &  1.00 $\pm$ 0.00 &  0.98 $\pm$ 0.01 &  1.00 $\pm$ 0.00 &  1.00 $\pm$ 0.00 &  4.74 $\pm$ 0.90 &  5.14 $\pm$ 0.81 &  5.14 $\pm$ 0.81
 \\  gconv3 &  1.40 $\pm$ 0.22 &  1.00 $\pm$ 0.00 &  1.00 $\pm$ 0.00 &  0.98 $\pm$ 0.01 &  1.00 $\pm$ 0.00 &  1.00 $\pm$ 0.00 &  5.75 $\pm$ 0.68 &  6.63 $\pm$ 0.54 &  6.63 $\pm$ 0.54
 \\ gconv4 &  1.00 $\pm$ 0.00 &  1.00 $\pm$ 0.00 &  1.00 $\pm$ 0.00 &  0.99 $\pm$ 0.00 &  1.00 $\pm$ 0.00 &  1.00 $\pm$ 0.00 &  5.66 $\pm$ 0.83 &  5.66 $\pm$ 0.83 &  5.66 $\pm$ 0.83

 \\  \cmidrule(l){1-1} \cmidrule(l){2-10} \textbf{Tiny 2}  &   \multicolumn{3}{c}{\textbf{Model Rank}} &  \multicolumn{3}{c}{\textbf{Corr}} & \multicolumn{3}{c}{\textbf{Testing Improvement}} 
\\ \textbf{Data}  &  \multicolumn{3}{c}{(--)} & \multicolumn{3}{c}{--} & \multicolumn{3}{c}{(\%)}
\\ $n_{models}=10$  & $g^{ph}$ & $g^{noph}$ & $g^{both}$ & $g^{ph}$ &  $g^{noph}$ & $g^{both}$ & $g^{ph}$ & $g^{noph}$ & $g^{both}$ 
\\ \cmidrule(l){1-1} \cmidrule(l){2-10} 
 tconv3 &  3.50 $\pm$ 0.32 &  2.65 $\pm$ 0.29 &  3.02 $\pm$ 0.33 &  0.47 $\pm$ 0.05 &  0.60 $\pm$ 0.04 &  0.62 $\pm$ 0.04 &  4.82 $\pm$ 0.85 &  6.58 $\pm$ 0.87 &  6.35 $\pm$ 0.92
\\ tconv4 &  3.25 $\pm$ 0.30 &  3.02 $\pm$ 0.31 &  2.50 $\pm$ 0.24 &  0.61 $\pm$ 0.05 &  0.71 $\pm$ 0.04 &  0.73 $\pm$ 0.04 &  7.72 $\pm$ 0.94 &  8.42 $\pm$ 1.08 &  9.67 $\pm$ 0.86

\\ tconv5 &  2.83 $\pm$ 0.31 &  2.80 $\pm$ 0.31 &  2.67 $\pm$ 0.30 &  0.58 $\pm$ 0.04 &  0.67 $\pm$ 0.04 &  0.68 $\pm$ 0.05 &  9.13 $\pm$ 0.92 &  8.87 $\pm$ 0.90 &  9.56 $\pm$ 0.91
\\  \midrule
\end{tabular}}
\caption{To evaluate our model $h''$  we determine the rank our current model selection in comparison to the others in terms of fine-tuning performance.  We also examine how our linear models correlate with the actual predicted testing performance. Lastly, we examine how our models compare to a randomly selected model in the bank of pre-trained models and record the average testing improvement compared to the randomly selected baseline. $n_{models}$ denotes the number of models used during the cross-validation procedure. This implies we could choose from $n_{models}-1$ models after each linear model $h''$ is constructed. The standard error is recorded to the right of each statistic.} 

\label{tab:task_sim_table}

\end{table*}

\subsection{Training Procedures}
\textbf{Conventionally Training Models:} For each classification problem we train 10 different parameter initializations of the same architecture using all training data provided. Each of the models are randomly initialized via the default initialization in Pytorch.
Training was performed with the Adam optimizer with mini-batch of 32 at each iteration and a learning rate of .01. All models were trained with a cross-entropy loss. We trained for 10, 10, and 50 epochs for the synthetic, greyscale, and tiny-ImageNet datasets respectively. We say that the DNN is learned if the training and testing accuracy is above a specific threshold. Specifics on the hyperparameters are defined in the supplementary material.

\textbf{Training with a Smaller Training Set:} To construct overfitted models, we train using a small number of randomly selected samples from the original training sets, keeping each class balanced. The loss function and optimization strategy is identical to the conventionally trained models.  We set the batch size equal to the number of training samples.  We say that a DNN has overfitted if the training accuracy and testing accuracy differ by a particular threshold. We recorded the training and setting performance at each step. Interestingly enough over-fitted models achieve close to 100\% training accuracy within the first 100 training steps. We apply this training procedure to 10 different parameter initializations for each architecture, each with its own unique subset of training samples. See supplementary material for specifics on the number of steps and the number of training samples used for each task.


\begin{figure*}[bp!]
\begin{center}
 \includegraphics[width=1.0\linewidth]{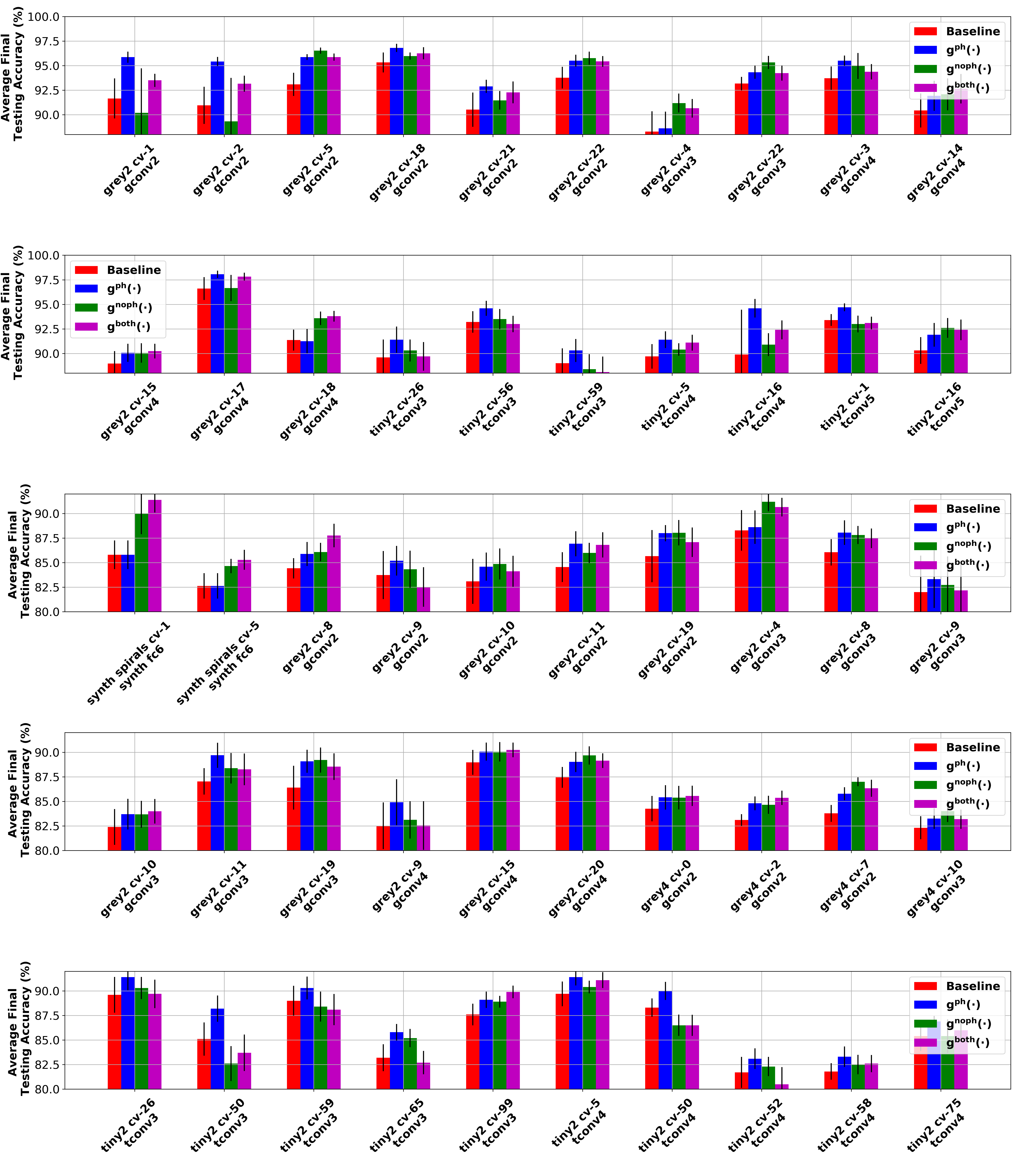}
\end{center}
  \caption{Shown above is the average final testing performance recorded after training (10 initialization for each model and task standard error denoted in black). We show there are classification problems where the topological meta-learning strategies consistently outperform the baseline training procedure denoted in red}
\label{fig:meta_1}
\end{figure*}

\begin{figure*}[bp!]
\begin{center}
 \includegraphics[width=1.0\linewidth]{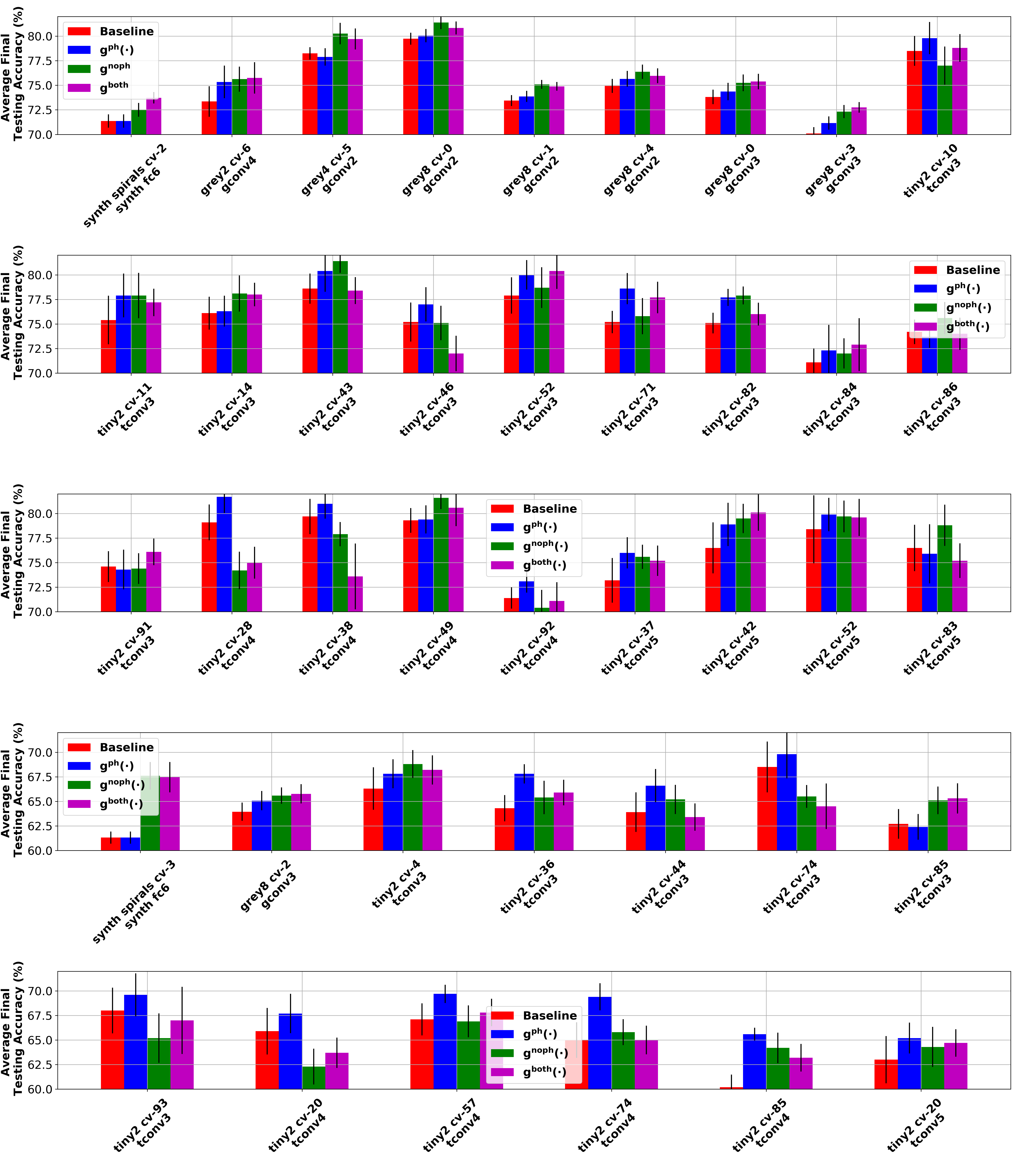}
\end{center}
  \caption{Additional results depict the effects of our meta-learning strategy on more challenging tasks. We similarly show there are classification problems where the topological meta-learning strategies consistently outperform the baseline training procedure denoted in red.}
\label{fig:meta_2}
\end{figure*}

\textbf{Fine-tuning Pre-trained Models for Task-Similarity:}
When fine-tuning models we apply the same training procedure as the models trained on the smaller training sets. The only distinction is that models were initialized using the parameters saved after the conventional training procedure from a different classification task. Recall we only select pre-trained parameters from models trained on tasks that differ from the current task being fine-tuned. To keep our model bank $\mathbfcal{F}_s$ small during cross-validation, we only consider one model initialization from each class. However, we fine-tune each model with 10 different small training sets. For the Tiny 2 Dataset, we select a subset of models to ensure the number of models needed for fine-tuning was attainable. See supplementary material for additional details.

\textbf{Topological Feature Generation For Performance Evaluation, Task-Similarity, and Meta-Learning:}   
To generate the topological features from each model for the performance estimation and meta-learning strategy, we pass the entire training set into the model. We then record the training and testing accuracy. We do this first before and after each training procedure. Additional details regarding our particular selection of nodes are defined in equations \ref{eq:16},\ref{eq:17} and \ref{eq:23} is located in our supplementary material.  All lasso models $h$,$h'$, and $h''$ are equipped with an alpha term of .01 and k=3 for each k-nearest neighbor models to infer model state.


\textbf{Training Models with Topological Constraints and Performance Evaluation:} 
 We use the same parameter initialization and sub-sampled training data as the DNNs that were prone to overfit. However, we train using our meta-learning strategy as defined in Section \ref{sec:meta}. We construct $\mathbf{T^*}$ by selecting topological features induced by DNNs of the same architecture conventionally trained \textbf{on other classification tasks} where testing performance indicates that the models were properly trained, i.e., the gap between training and testing accuracy is small and the testing accuracy is above a particular threshold. 


Models are evaluated after each training step. Models are evaluated using the same sequestered testing dataset until training accuracy remains constant. This usually occurs between 100 to 250 steps. For comparison, we compared the final recorded testing accuracy throughout training for the case with and without the topological regularizer. Figures  \ref{fig:meta_1} and \ref{fig:meta_2} shows examples of some of the partitions of the data for synthetic, greyscale, and tiny-ImageNet datasets. Additional details about the meta-learning procedure are found in the supplementary material.


\section{Results}

\textbf{Performance Estimation Results:}
We evaluate our performance estimation via $|C|$-fold cross-validation where $|C|$ denotes the number of classification tasks in each parent class. For the intended model state estimation we compute the average classification accuracy across tasks for our nearest neighbor models $\mathbb{N}\mathbb{N}_k$. The performance of the linear models $h$ and $h'$ used to estimate the testing performance and performance gap are evaluated using the mean absolute error between estimated values and true values. Table \ref{tab:performance_tab} exhibits our ability to estimate the performance of a given model. We can relatively easily infer between model states even with features that do not require persistent homology analysis.  Our testing accuracy \textit{prediction} is under 10.0\% overall in terms of mean absolute error. We observed a similar trend in  performance gap estimation. and find that out performance gap estimation is under 7.0\% in terms of mean absolute error.

\textbf{Task-Similarity}
To evaluate task similarity we consider how well $h''$ was in selecting the most appropriate model for fine-tuning. We accomplish this by ranking our model selection with respect to all of the other models that could have been selected according to their fine-tuning performance on the current tasks. Additionally, we compute the correlation between the estimated fine-tuning performance from $h''$ with the actual fine-tuning performance for each task.  Lastly, we compare the selected fine-tuning accuracy in comparison to the average fine-tuning accuracy of the DNNs being fine-tuned. 
Similar to the previously defined performance estimation strategy, we applied this method in a leave one task out cross-validation procedure, meaning that we trained models using models trained on every task but task $\mathcal{T}_c$. Results are shown and described in Table \ref{tab:task_sim_table}. We can see this simple linear model is effective in selecting appropriate models as we consistently achieve a better testing performance than the expected testing performance of a randomly selected model for fine-tuning. 

\textbf{Meta-Learning:}
We examine how our topological meta-learning strategy impacts the testing performance of models being trained on smaller training sets.  As previously defined in the main manuscript we quantify performance by examining the final recorded testing accuracy after training. In Figures \ref{fig:meta_1} and \ref{fig:meta_2} we show a set of sample classification problems where our meta-learning optimization strategy improves testing performance. Baseline training is denoted in red while the other colors refer to various topological characterizations with the same hyperparameter selection. Error bars are a result of multiple trials using different initializations and training data samples. We find that there are many classification problems that benefit consistently from our meta-learning strategy. Moreover, this strategy appears to be consistently successful across various architectures tackling the same task. 

\section{Discussion \& Conclusion}
  We consider further avenues of exploration for the topological analysis of deep networks. One possible avenue is to more deeply understand the meaning of each of these topological features that are useful for performance estimation. One interpretation of previous works which use persistent homology of deep networks is that it can be used to quantify the sparsity of the computational graph in the DNN. This is consistent with previous works like ``the lottery ticket hypothesis"  \cite{frankle2018lottery} which claims there exists a sub-graph within DNNs that is doing the majority of the computational work. Though our characterization was inspired by previous topological characterizations, the interpretation of model sparsity is not so explicit.  Examining how our features correlate with previous topological characterizations may be of interest. Additionally, we hope this work inspires others to characterize the structure of DNNs while being less computationally expensive.

Another avenue would be to construct topological characterizations which are invariant to architecture. Since many of our topological features consider local structures of the DNN, it would be possible to compare components of a given network with another component of a network, even if both models consist of different architectures. Lastly, topological characterizations for novel architectures like transformers should be explored.

In summary, we proposed a practical framework to topologically characterize fully connected and convolutional layers in DNNs. Our proposed topological features are quick to compute, differentiable, and can be used in a variety of applications. We showed that the topological characterization can be used to estimate the progress of learning. We also showed its use in examining whether a model was appropriate for fine-tuning on a new task.  In addition, we proposed a meta-learning strategy that forces DNNs to elicit topological features consistent with learning. We show that this topological regularization can improve testing performance across numerous datasets and architectures.

\ifCLASSOPTIONcompsoc
  \section*{Acknowledgments}
\else
  \section*{Acknowledgment}
\fi

This research was supported by the
National Institutes of Health (NIH), grants R01-DC-014498
and R01-EY-020834, the Human Frontier Science Program
(HFSP), grant RGP0036/2016, The Ohio Super Computing Center and a grant from Ohio State’s
Center for Cognitive and Brain Sciences

\ifCLASSOPTIONcaptionsoff
  \newpage
\fi

\begin{IEEEbiography}[{\includegraphics[width=1in,height=1.25in,clip,keepaspectratio]{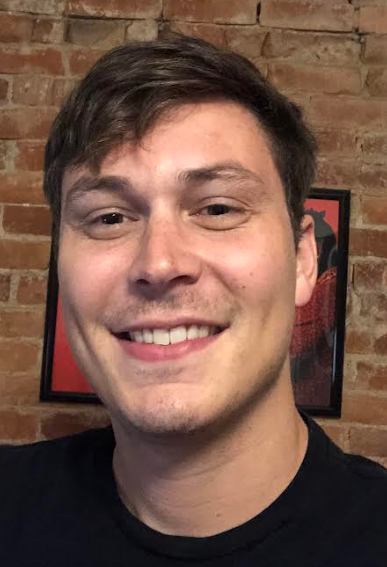}}]{Stuart Synakowski}
received bachelors degrees in Physics and Applied Mathematics/Statistics at Clarkson University, Potsdam, New York in 2017. During his undergraduate studies, he partook in computational biophysics research.
He is currently a Ph.D. candidate in Electrical and Computer Engineering at The Ohio State University, working as a research assistant within the Computational Biology and Cognitive Science Lab. His current research interests are in topological data analysis (TDA) applied to improving the performance and understanding of deep neural networks.  He is also interested in developing computer vision systems equipped with higher-level action understanding of agents.
\end{IEEEbiography}

\begin{IEEEbiography}[{\includegraphics[width=1in,height=1.25in,clip,keepaspectratio]{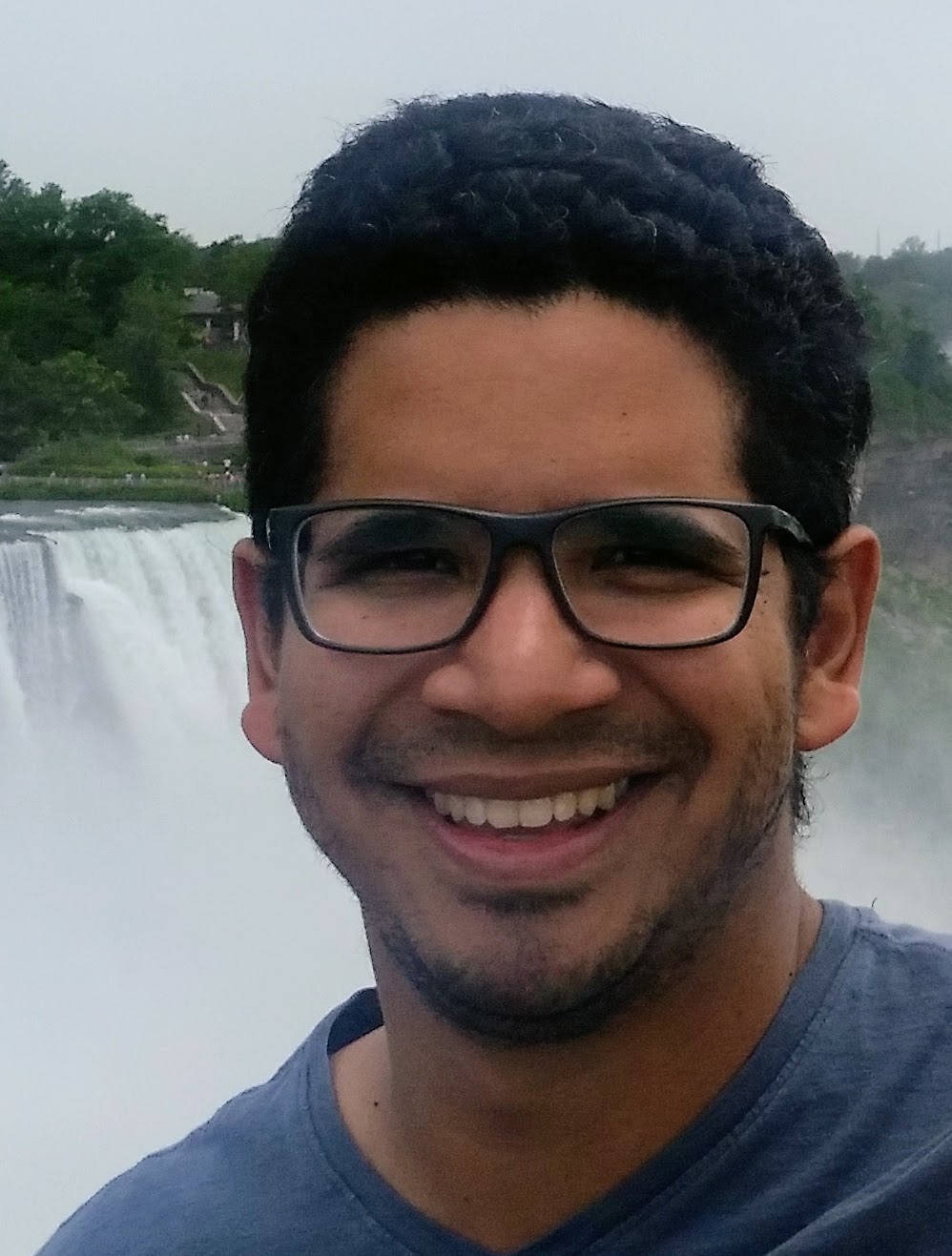}}]{Fabian Benitez-Quiroz}
received the BS degree in electrical engineer-
ing from Pontificia Universidad Javeriana, Cali,
Colombia, in 2004, the MS degree in electrical
engineering from the University of Puerto Rico,
Mayaguez, Puerto Rico, in 2008, and the PhD
degree in electrical and computer engineering
from the Ohio State University (OSU), in 2015.
He is currently a postdoctoral researcher with the
Computational Biology and Cognitive Science
Lab at OSU. His current research interests
include the analysis of facial expressions in the wild, functional data
analysis, deformable shape detection, face perception, and deep learning
. He is a member of the IEEE.
\end{IEEEbiography}

\begin{IEEEbiography}[{\includegraphics[width=1in,height=1.25in,clip,keepaspectratio]{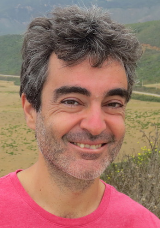}}]{Aleix M. Martinez}
Aleix M. Martinez is a professor with the Depart-
ment of Electrical and Computer Engineering,
The Ohio State University (OSU), where he is
the founder and director of the Computational
Biology and Cognitive Science Lab. He is also
affiliated with the Department of Biomedical Engi-
neering and to the Center for Cognitive and Brain
Sciences, where he is a member of the executive
committee. He has served as an associate editor
of the IEEE Transactions on Pattern Analysis and
Machine Intelligence, the IEEE Transactions on
Affective Computing, Image and Vision Computing, and Computer
Vision and Image Understanding. He has been an area chair for many
top conferences and was program chair for CVPR, 2014. He is also a
member of NIH’s Cognition and Perception study section. He is most
known for being the first to define many problems and solutions in face
recognition (e.g., recognition under occlusions, expression, imprecise
landmark detection), discriminant analysis (e.g., Bayes optimal solu-
tions, subclass-approaches, optimal kernels), structure from motion
(e.g., using kernel mappings to better model non-rigid deformations,
noise invariance), and, most recently, demonstrating the existence of a
much larger set of cross-cultural facial expressions of emotion than
previously known (i.e., compound expressions of emotion) and the
transmission of emotion through changes in facial color
\end{IEEEbiography}





\end{document}


%
\title{Leveraging The Topological Consistencies of Learning in Deep Neural Networks: \\ Supplementary Material}
%
%
%
%

\author{Stuart~Synakowski,Fabian Benitez-Quiroz,
        Aleix Martinez 
\IEEEcompsocitemizethanks{\IEEEcompsocthanksitem S. Synakowski was with the Department
of Electrical and Computer Engineering, The Ohio State University, Columbus,
OH, 43210.\protect\\
E-mail: synakowski.1@buckeyemail.osu.edu
}
\thanks{Manuscript received ---; revised ---}}

%
%

\markboth{Under Review October 2021}%
{Shell \MakeLowercase{\textit{et al.}}: Bare Demo of IEEEtran.cls for Computer Society Journals}
%



\IEEEtitleabstractindextext{%
\begin{abstract}
Many works have tried to determine the properties of deep neural networks (DNNs) that ensure good performance on a given task. Recently, methods have been developed to accurately predict a DNN's testing performance on a particular task given statistics of its underlying topological structure. However, further leveraging this newly found insight for practical applications is intractable due to the high computational cost in terms of time and memory. In this work we define a new class of \textit{topological features} that accurately characterize the progress of learning, while being quick to compute during running time. Moreover, our proposed topological features are readily equipped for backpropagation, meaning that they can be incorporated in end-to-end training. Our proposed features can be used in numerous applications. Not only can we accurately estimate the performance of DNN across tasks in real-time but we can induce learning by constraining the topological structure of networks.
\end{abstract}

\begin{IEEEkeywords}
Topological Data Analysis, Deep Learning, Explainable AI.
\end{IEEEkeywords}}



\IEEEdisplaynontitleabstractindextext

\IEEEpeerreviewmaketitle

%


%

\section{Topological Characterization of Convolutional Layers}
We begin by considering a hidden feature representation of $\mathbf{n}$ samples with a dimension greater than two. 
We define $\mathbf{h}^{conv}_i$ as the hidden feature representation of $n$ samples before being convolved by the $i$-th layer as 

\begin{equation}
\mathbf{h}^{conv}_i \in \mathbb{R}^{\mathbf{n} \times |\tilde{\mathbf{c}}_i|\times |\tilde{\mathbf{l}}_i|  \times |\tilde{\mathbf{w}}_i| }, 
\end{equation}
Where $\mathbf{n}$, $|\tilde{\mathbf{c}}_i|$,$|\tilde{\mathbf{l}}_i|$,$|\tilde{\mathbf{w}}_i|$ denote the number of samples, number of channels, the tensor height, and tensor width respectively. 
For indexing purposes we define
\begin{equation}
 \mathbf{h}^{conv}_{i,\cdot, \tilde{c}_i,\tilde{l}_i  \tilde{w}_i } \subset \mathbf{h}^{conv}_i
\end{equation}
as the set of $\mathbf{n}$ features in $\mathbb{R}^\mathbf{n}$ corresponding to the $i-th$ layer of the $\tilde{c}_i$-th  channel $\tilde{l}_i $-th row, and $\tilde{w}_i $-th column. We define a convolutional layer as
\begin{equation}
\mathbf{h}^{conv}_{i+1}=  \phi_{\text{ReLU}}( W^{conv}_i \mathbf{\ast} \mathbf{h}^{conv}_i   \oplus \mathbf{b}_i^{conv} )
\end{equation}where $W^{conv}_i \in \mathbb{R}^{|\tilde{\mathbf{c}}_{i+1}|\times |\tilde{\mathbf{c}}_i|\times|\tilde{\mathbf{k}}_i|\times|\tilde{\mathbf{k}}_i| }$, such that $|\tilde{\mathbf{k}}_i|$ denotes the filter size. $\phi_{ReLU}$ is an element-wise activation function.
We define $\mathbf{\ast}$ as an operation that computes a set of convolutions across the height and width of the tensor with zero padding and a stride of 1 across $\mathbf{h}^{conv}_{i}$, and $\oplus^{conv} $ adds a bias term $\mathbf{b}_i^{conv} \in \mathbb{R}^{|\tilde{\mathbf{c}}_{i+1}|}$  corresponding to each channel. Hence for each element in $\mathbf{h}^{conv}_{i+1}$
\begin{multline}
 \mathbf{h}^{conv}_{i+1, \tilde{c}_{i+1},\tilde{l}_{i+1}  \tilde{w}_{i+1} }  = \\  \phi_{ReLU} \Big(  \sum_{\tilde{c}_i=1}^{|\tilde{\mathbf{c}}_{i}|}  \sum_{\tilde{l}_i= \tilde{l}_{i+1} }^{\tilde{l}_{i+1}+|\tilde{\mathbf{k}}_i|}      \sum_{\tilde{w}_i= \tilde{w}_{i+1} }^{\tilde{w}_{i+1}+|\tilde{\mathbf{k}}_i|}    W^{conv}_{i,\tilde{c}_{i},\tilde{l}_i.\tilde{w}_i}  \mathbf{h}^{conv}_{i,\cdot, \tilde{c}_i,\tilde{l}_i  \tilde{w}_i }  + b_{i,\tilde{c}_{i}} \Big).
 \end{multline}

We also define the set of statistics on the set of hidden feature representations being fed into the convolutional layer. We define $\mathbf{h}^{conv,\mu}_i \in \mathbb{R}^{|\tilde{\mathbf{c}}_i|\times|\tilde{\mathbf{k}}_i|\times|\tilde{\mathbf{k}}_i|},$ which is the set of mean activations being applied to each element of each filter. For each element in $\mathbf{h}^{conv,\mu}_i$,
\begin{multline}
 \mathbf{h}^{conv,\mu}_{ i,\tilde{c}_{i},\tilde{l}_{i} , \tilde{w}_{i} }  = \frac{1}{|\tilde{\mathbf{k}}_i||\tilde{\mathbf{k}}_i|\mathbf{n}}\sum_{n_i=1}^{\mathbf{n}}   \sum_{l_i= \tilde{l}_{i} }^{\tilde{l}_{i}+|\tilde{\mathbf{l}}_i|-|\tilde{\mathbf{k}}_i|}     \sum_{w_i= \tilde{w}_{i} }^{\tilde{w}_{i}+|\tilde{\mathbf{w}}_i|-|\tilde{\mathbf{k}}_i|}     \mathbf{h}^{conv}_{i,n_i, \tilde{c}_i,l_i ,w_i }  .
\end{multline}
We also perform the same set of hidden activations to construct the standard deviation in activations $\mathbf{h}^{conv,\sigma}_{ i,\tilde{c}_{i},\tilde{l}_{i} , \tilde{w}_{i} }$.

Additionally, we define the set of mean activations for each channel for a given sample as $\mathbf{h}^{conv,\mu'}_i \in \mathbb{R}^{ \mathbf{n} \times |\tilde{\mathbf{c}}_i|}$, where each vector in $\mathbf{h}^{conv,\mu'}_i $ is defined by   

\begin{equation}
\mathbf{h}^{conv,\mu'}_{i,\cdot,c_i}=\frac{1}{|\tilde{\mathbf{l}}_i||\tilde{\mathbf{w}}_i|}\sum_{\tilde{l}_i=1}^{|\tilde{\mathbf{l}}_i|} \sum_{ \tilde{w}_i=1}^{|\tilde{\mathbf{w}}_i|} h^{conv}_{i,\cdot,\tilde{c}_i,\tilde{l}_i  \tilde{w}_i} 
\end{equation}We additionally define $\mathbf{h}^{conv, \mu''}_{i,\cdot,c_i}  \in \mathbb{R}^{ |\tilde{\mathbf{c}}_i|}$as the set of mean-channel wise activations for each sample in $\mathbf{h}^{conv,\mu'}_{i,\cdot,c_i}$. Similarly, we define $\mathbf{h}^{conv, \sigma''}_{i,c_i}  \in \mathbb{R}^{ |\tilde{\mathbf{c}}_i|}$ as the set of standard deviation values in activations for each sample in $\mathbf{h}^{conv,\mu'}_{i,c_i}$.

\subsection{By Filter}
Similar to our topological features induced by a single node in a fully connected layer, we define the following 1-dimensional point sets for each $c_i$-th filter in the $i$-th layer as
\begin{equation}
    A^{conv}_{i,\tilde{c}_i} = \{W^{conv}_{i,\tilde{c}_{i},\tilde{l}_i.\tilde{w}_i} \mathbf{h}^{conv,\mu}_{i,\tilde{c}_i,\tilde{l}_i ,\tilde{w}_i}| \tilde{l}_i \in 1 ... |\tilde{\mathbf{l}}_i| \text{ and } \tilde{w}_i  \in  1... |\tilde{\mathbf{w}}_i|  \}.
\end{equation}
Similarly, we define the following point sets by computing the standard deviation in activations
\begin{equation}
    I^{conv}_{i,\tilde{c}_i} = \{\|W^{conv}_{i,\tilde{c}_{i},\tilde{l}_i.\tilde{w}_i} \mathbf{h}^{conv,\sigma}_{i,\tilde{c}_i,\tilde{l}_i ,\tilde{w}_i}|| \tilde{l}_i \in 1 ... |\tilde{\mathbf{l}}_i| \text{ and } \tilde{w}_i  \in  1... |\tilde{\mathbf{w}}_i|\}.
\end{equation}We then compute the topological characterization that was computed for sets of nodes in fully-connected layers using our previously defined topological characterization operation $\mathbb{T}(\cdot)$.

\begin{equation}
\mathbb{T}(\mathbf{A}^{conv}_i) = \{\mathbf{g}(A^{conv}_{i,1})),\mathbf{g}(A^{conv}_{i,2}) ...\}
\end{equation}\begin{equation}
\mathbb{T}(\mathbf{I}^{conv}_i) = \{\mathbf{g}(I^{conv}_{i,,1})),\mathbf{g}(I^{conv}_{i,2}) ...\}
\end{equation}
\subsection{By Layer}
To construct a topological representation for a given convolutional layer, we collapse the hidden representation of $\mathbf{h}^{conv}_i $ by computing the mean activation values across the height and width of the feature representation and then compute statistics on the remaining feature presentation. Hence we define
\begin{equation}
H^{conv}_{\mu_i} = \{\mathbf{h}^{conv, \mu''}_{i,c_i}   | \tilde{c}_i  \in  1... |\tilde{\mathbf{c}}_i| \}
\end{equation}
and
\begin{equation}
H^{conv}_{\sigma_i} = \{\mathbf{h}^{conv, \sigma''}_{i,c_i} | \tilde{c}_i  \in  1... |\tilde{\mathbf{c}}_i| \}
\end{equation}with the final topological characterization for the layer as
\begin{equation}
t^{Hconv}_i = [\mathbf{g}(H^{conv}_{\mu_i}),\mathbf{g}(H^{conv}_{\sigma_i}) ].
\end{equation}
\subsection{By Architecture}

Lastly, we compute correlation in activation between the mean activation values corresponding to $\mathbf{h}^{conv,\mu'}_{i,\cdot,c_i}$and the activations values corresponding to nodes throughout the network. Hence
\begin{equation}
C^{conv }_{i,\tilde{c}_i}=  \{cov(\mathbf{h}^{conv,\mu'}_{i,\cdot,c_i},\mathbf{h}_{k}) | k \in \mathcal{C}_{ij}^{\text{ind}}\},
\end{equation}where $\mathcal{C}_{ij}^{\text{ind}}$ denotes indices corresponding  within convolutional or fully-connected layers. We than collect sets of these point sets to construct a set of topological representations.
\begin{equation}
\mathbb{T}(\mathbf{C}_i^{conv}) = \{\mathbf{g} (C^{conv}_{i,1}),\mathbf{g}(C^{conv}_{i,2}) ...\}.
\end{equation}
\subsection{Aggregating Features}
Similar to our topological feature representation of fully-connected layers we then aggregate these features into one feature vector $\mathbf{t}^{conv}_i$

where
\begin{multline}
\mathbf{t}_i^{conv}=[g'(\mathbb{T}(\mathbf{A}^{conv}_i)),
g'(\mathbb{T}(\mathbf{I}^{conv}_i)), \\  
g'(\mathbb{T}(\mathbf{C}^{conv}_i)),t^{Hconv}_i ],
\end{multline} using $g'(\cdot)$ which was defined in the methods section.

\section{Additional Topological Construction Details}
\textbf{Random Point-Set Selection:} For the point-sets $A''_{il}$ and $I''_{il}$ defined in equations 16 and 17 in the main manuscript, we randomly select 10 sets of 10 nodes from each hidden fully connected layer for all models defined in our experiments. From the 10 sets of 10 nodes, we apply our topological characterization $\mathbf{g}$ and then $g'$ to make the topological characterization.

\textbf{Selection and Grouping of Covariance Between Nodes:}
We explore two characterizations to topologically characterize the covariance in activations between nodes. We first consider the covariance in activations between each of the nodes in the hidden layers with the output activations. A single 1-d point-set is defined by examining the covariance between nodes in a single layer and a single output node. Assuming there are $\mathbf{c}$ classes we would have $\mathbf{c}$ covariance point-sets corresponding to each layer. We then apply $\mathbf{g}$ and $g'$ to construct the feature representation for the given layer.
For the second covariance characterization, we consider the covariance between a single node in a hidden layer and all of the other nodes in the entire network. This large point-set is then characterized by $\mathbf{g}$ and then grouped with all of the other covariance induced point-sets in the same layer before being operated on by $g'$.

\textbf{Required Down-Sampling of Degenerate Points}
One of the requirements for the topological optimization strategy defined in \cite{bruel2019topology} is that the inverse map between simplices and point-sets must be unique. The means if there are any duplicate points in the construction of a point-set we will be unable to optimize for particular persistent homology features. To address this issue we randomly select a point if there are any duplicate points during the point-set constructions.
\section{Data Set Partitioning Details}
Tables \ref{tab:synth_data} \ref{tab:grey2_class_names} \ref{tab:grey4_class_names} \ref{tab:tiny2_class_names} Present the particular partitioning of classification problems for our experiments.

\section{Additional Training Details}
We recorded the hyper-parameters in table \ref{tab:conv_traing_details} for the conventional training procedures.
\begin{table}[h]
\begin{center}
\begin{adjustbox}{max width=\textwidth}
\scalebox{0.85}{
\begin{tabular}{|l|c|}
\hline
Parameters  & Values \\
\hline\hline
Conventional Loss Term  &  Cross-Entropy\\ \hline 
Optimizer  & Adam \\ \hline
Learning Rate  & .01 \\ \hline
beta terms & .9,.999 \\ \hline
batch size  & 32      \\ \hline
number of epochs synthetic &  10 \\ \hline
number of epochs grey2 &  10 \\ \hline
number of epochs grey4 &  10 \\ \hline
number of epochs grey8 &  10 \\ \hline
number of epochs tiny2 &  50 \\ \hline
\end{tabular}}
 \end{adjustbox}
\end{center}
\caption{Parameters for training models conventionally}
\label{tab:conv_traing_details}
\end{table}

\section{Overfitting Models}
See table \ref{tab:overfitting_details} for details on overfitting procedure such as the optimization strategy the number of training samples used and the number of training steps.
\begin{table}[h]
\begin{center}
\begin{adjustbox}{max width=\textwidth}
\scalebox{0.85}{
\begin{tabular}{|l|c|}
\hline
Parameters  & Values \\
\hline\hline
Conventional Loss Term  &  Cross-Entropy\\ \hline 
Optimizer  & Adam \\ \hline
Learning Rate  & .01 \\ \hline
beta terms & .9,.999 \\ \hline
batch size  & all      \\ \hline
number of training samples spirals (each class)  & 25\\\hline
number of training samples (each class)  & 10        \\\hline
number of training samples xor (each class)  & 8     \\\hline
number of training samples circles (each class)  & 10\\\hline
number of training samples gauss (each class)  & 2   \\ \hline
number of training samples grey2 (each class)  & 10  \\ \hline
number of training samples grey4 (each class)  & 10  \\ \hline
number of training samples grey8 (each class)  & 10  \\ \hline
number of training samples tiny2 (each class)  & 50  \\ \hline
number of steps synthetic &  250 \\ \hline
number of steps grey2 &  100 \\ \hline
number of steps grey4 &  100 \\ \hline
number of steps grey8 &  100 \\ \hline
number of steps tiny2 &  100 \\ \hline

\end{tabular}}  
 \end{adjustbox}
\end{center}
\caption{Parameters for overfitting models}
\label{tab:overfitting_details}
\end{table}

\section{Fine-Tuning Additional Details} We apply the same overfitting training procedure but using pre-trained models that were trained using the conventional training procedure. For the toy and greyscale data sets we apply cross-validation of every classification problem in the parent data set. Since we did not have the computational resources to fine-tune 300k+ models on the tiny2 task similarity tasks, we selected a subsample of 4 groups of 10 tasks and apply 10 fold cross-validation on each subset. We select the subset of tasks in the tiny2 parent data set if the difference in conventional training and overfitting testing performance differed by 15\%.

\begin{table}[h]
\begin{center}
\begin{adjustbox}{max width=\textwidth}
\scalebox{0.85}{
\begin{tabular}{|l|c|}
\hline
group  & cv-indices \\
\hline\hline

group1 & 8, 9, 11, 13, 14, 19, 21, 23, 24, 27\\ \hline
group1 & 28, 39, 43, 45, 46, 48, 49,50, 51, 52 \\ \hline
group1 & 53, 55, 58, 61, 64, 65, 66, 71, 74, 76 \\ \hline
group1 & 77, 81, 82, 83, 84, 87, 88, 89, 94, 98 \\ \hline
\end{tabular}}
 \end{adjustbox}
\end{center}
\caption{Grouping of tiny2 classification problems for the task-similarity cross-validation experiments. We apply the cross-validation procedure across each class within each group and then aggregate all results across classification problems and groups for the final results.}
\label{tab:tiny_2_fine_tuning}
\end{table}

\section{Training Data Construction For Lasso Based Performance Estimation}
To estimate the performance testing performance of models we select a subset of topological features where the observable training performance was above a particular threshold. table \ref{tab:lasso_thresholds} shows these thresholds.
\begin{table}[h]
\begin{center}
\begin{adjustbox}{max width=\textwidth}
\scalebox{0.85}{
\begin{tabular}{|l|c|}
\hline
parent class  & training threshold (accuracy \%) \\
\hline\hline

synthetic 2D & 98\%    \\ \hline
grey2 & 75\%    \\ \hline
grey4 & 50\%    \\ \hline
grey8 & 50\%    \\ \hline
tiny2 & 75\%    \\ \hline
\end{tabular}}
 \end{adjustbox}
\end{center}
\caption{Training thresholds for the linear models used for performance estimation}
\label{tab:lasso_thresholds}
\end{table}

\section{Additional Meta-Leaning Parameters} 
We have defined the set of required hyper-parameters needed for our meta-learning optimization strategy. To save on  we only restrict ourselves to optimizing the hidden activation results described in equations 20 and 21 of the manuscript. See table \ref{tab:meta-learning}
for the list of all parameters needed for the meta-learning strategies.
\section{Hyper-parameters Selected For Meta-Learning}
\begin{table}[h]
\begin{center}
\begin{adjustbox}{max width=\textwidth}
\scalebox{0.85}{
\begin{tabular}{|l|c|}
\hline
Parameters  & Values \\
\hline\hline

Conventional Loss Term  &  Cross-Entropy\\ \hline 
Lambda Term $\lambda$ & 0.05 \\ \hline
Optimizer  & Adam \\ \hline
Learning Rate  & .01 \\ \hline
beta terms & .9,.999 \\ \hline
train samples grey2 (each class) & 10 \\ \hline
train samples grey4 (each class) & 10 \\ \hline
train samples grey8 (each class) & 10 \\ \hline
train samples tiny2 (each class) & 50 \\ \hline
number of  training steps &  100 \\ \hline
random topological selection at each step & 25 \\ \hline
$\text{min}_k$ &  5 \\ \hline
correlation threshold synthetic  & 0.6 \\ \hline
correlation threshold grey2  & 0.6 \\ \hline
correlation threshold grey4  & 0.6 \\ \hline
correlation threshold grey8  & 0.6 \\ \hline
correlation threshold tiny2  & 0.5 \\ \hline
test threshold synthetic 2d  & 99\% \\ \hline
test threshold grey2  & 99\% \\ \hline
test threshold grey4  & 97\% \\ \hline
test threshold grey8  & 95\% \\ \hline
test threshold tiny2  & 90\% \\ \hline
pergap threshold synthetic 2d  & 2\% \\ \hline
pergap threshold grey2  & 3\% \\ \hline
pergap threshold grey4  & 8\% \\ \hline
pergap threshold grey8  & 15\% \\ \hline
pergap threshold tiny2  & 10\% \\ \hline
\end{tabular}}
 \end{adjustbox}
\end{center}
\caption{Hyper-parameters for the topological meta-learning strategy. $\lambda$ is for scaling the topological term in the loss function. Learning rate and beta terms are the hyper-parameters for the Adam optimizer\cite{kingma2017adam}. 
$\mathbf{T^*}$ was selected from a bank of topological features where the testing performance was above the test threshold and the performance gap was below the performance gap threshold. During each cross-validation procedure $\mathbf{T^*}$ never includes topological features corresponding to the current classification problem. 
Training samples refer to the number of randomly selected training samples from each class used to train the model.  Random topological selection refers to the number of randomly selected topological features selected from $\mathbf{T}^*$. We then compute the weighted distance between each of the topological features and our current topological features using  the correlation threshold. Once distances are computed between current topological features and the randomly select topological features, the average of the closet $\text{min}_k$ features are used for the topological loss term.}
\label{tab:meta-learning}
\end{table}

\section{Running-Times}
We have recorded the average running times for each forward pass of the topological characterization for each of the models see table \ref{tab:running_times}. For simplicity, we group the topological characterizations by type of point-sets that were extracted from local, layer, and global characterizations.

\section{Computational Resources}
Any training procedure applied to any model or data set can be performed using 3 cores of an Intel Xeon E5-2680 V4 with 12 GB of RAM. A single training procedure of a single network on a particular task can be achieved in less than 10 minutes for all models and tasks.



\begin{table}[h] 
\centering 
\scalebox{1.0}{
\begin{tabular}{|l | c | c | }
 \hline \textbf{{Parent Class}}  &  \textbf{class and subclass} & \textbf{augmentation}  
 \\ \hline \hline synthetic 2d & spirals 0 & rotate $0^{\circ}$ scale-x 1
 \\ \hline synthetic 2d & spirals 1 & rotate $45^{\circ}$ scale-x 1
 \\ \hline synthetic 2d & spirals 2 & rotate $9^{\circ}$ scalex 1
 \\ \hline synthetic 2d & spirals 3 & rotate $0^{\circ}$ scalex 2
 \\ \hline synthetic 2d & spirals 4 & rotate $45^{\circ}$ scalex 2
 \\ \hline synthetic 2d & spirals 5 & rotate $90^{\circ}$ scalex 2
 \\ \hline synthetic 2d & moons 0 & rotate $0^{\circ}$ scale-x 1
 \\ \hline synthetic 2d & moons 1 & rotate $45^{\circ}$ scale-x 1
 \\ \hline synthetic 2d & moons 2 & rotate $9^{\circ}$ scalex 1
 \\ \hline synthetic 2d & moons 3 & rotate $0^{\circ}$ scalex 2
 \\ \hline synthetic 2d & moons 4 & rotate $45^{\circ}$ scalex 2
 \\ \hline synthetic 2d & moons 5 & rotate $90^{\circ}$ scalex 2
  \\ \hline synthetic 2d & xor 0 & rotate $0^{\circ}$ scale-x 1
 \\ \hline synthetic 2d & xor 1 & rotate $45^{\circ}$ scale-x 1
 \\ \hline synthetic 2d & xor 2 & rotate $9^{\circ}$ scalex 1
 \\ \hline synthetic 2d & xor 3 & rotate $0^{\circ}$ scalex 2
 \\ \hline synthetic 2d & xor 4 & rotate $45^{\circ}$ scalex 2
 \\ \hline synthetic 2d & xor 5 & rotate $90^{\circ}$ scalex 2
  \\ \hline synthetic 2d & gauss 0 & rotate $0^{\circ}$ scale-x 1
 \\ \hline synthetic 2d & gauss  1 & rotate $45^{\circ}$ scale-x 1
 \\ \hline synthetic 2d & gauss  2 & rotate $9^{\circ}$ scalex 1
 \\ \hline synthetic 2d & gauss  3 & rotate $0^{\circ}$ scalex 2
 \\ \hline synthetic 2d & gauss  4 & rotate $45^{\circ}$ scalex 2
 \\ \hline synthetic 2d & gauss  5 & rotate $90^{\circ}$ scalex 2
  \\ \hline synthetic 2d & circles 0 & rotate $0^{\circ}$ scale-x 1
 \\ \hline synthetic 2d & circles 1 & rotate $45^{\circ}$ scale-x 1
 \\ \hline synthetic 2d & circles 2 & rotate $9^{\circ}$ scalex 1
 \\ \hline synthetic 2d & circles 3 & rotate $0^{\circ}$ scalex 2
 \\ \hline synthetic 2d & circles 4 & rotate $45^{\circ}$ scalex 2
 \\ \hline synthetic 2d & circles 5 & rotate $90^{\circ}$ scalex 2
\\  \midrule
\end{tabular}}
\caption{Augmentation operations for the synthetic 2d problems. We denotes the scale factor of the non-isotropic scaling in the x component and number of degrees the data is rotated.   Scaling is applied before rotation. } 

\label{tab:synth_data}

\end{table}

\begin{table}[h] 
\centering 
\scalebox{1.0}{
\begin{tabular}{|l | c | c | }
 \hline \textbf{{Parent Class}}  &  \textbf{cv-index} & \textbf{Classes}  
 \\ \hline \hline grey2 & 0 & 5, m 
 \\ \hline grey2 & 1 & 4, 8
 \\ \hline grey2 & 2 & 0, a
 \\ \hline grey2 & 3 & 1, k1
 \\ \hline grey2 & 4 & f, i
 \\ \hline grey2 & 5 & e, x
 \\ \hline grey2 & 6 & k8, k9
 \\ \hline grey2 & 7 & k4, w
 \\ \hline grey2 & 8 & j, s
 \\ \hline grey2 & 9 & k5, d
 \\ \hline grey2 & 10 & k3, g
 \\ \hline grey2 & 11 & k, l
 \\ \hline grey2 & 12 & 2, p
 \\ \hline grey2 & 13 & u, v
 \\ \hline grey2 & 14 & 9, k2
 \\ \hline grey2 & 15 & n, y
 \\ \hline grey2 & 16 & q, r
 \\ \hline grey2 & 17 & 3, o
 \\ \hline grey2 & 18 & 7, z
 \\ \hline grey2 & 19 & k7, h
 \\ \hline grey2 & 20 & k0, b
 \\ \hline grey2 & 21 & k6, c
 \\ \hline grey2 & 22 & 6, t
\\  \midrule
\end{tabular}}
\caption{Partitioning of Data Sets for the grey2 classification problems. k\# denote the classes from the kmnist dataset. } 

\label{tab:grey2_class_names}

\end{table}

\begin{table}[h] 
\centering 
\scalebox{1.0}{
\begin{tabular}{|l | c | c | }
 \hline \textbf{{Parent Class}}  &  \textbf{cv-index} & \textbf{Classes}  
 \\ \hline \hline grey4 & 1 & 5, m, 4, 8
 \\ \hline grey4 & 1 & 0, a, 1, k1
 \\ \hline grey4 & 2 & f, i, e, x
 \\ \hline grey4 & 3 & k8, k9 , k4, w
 \\ \hline grey4 & 4 & j, s , k5, d
 \\ \hline grey4 & 5 & k3, g , k, l
 \\ \hline grey4 & 6 & 2, p, u, v
 \\ \hline grey2 & 7 & 9, k2, n, y
 \\ \hline grey2 & 8 & q, r, 3, o
 \\ \hline grey2 & 9 & 7, z , k7, h
 \\ \hline grey2 & 10 & k0, b , k6, c
\\  \midrule
\end{tabular}}
\caption{ Partitioning of Data Sets for the grey4 classification problems. k\# denote the classes from the kmnist dataset. } 
\label{tab:grey4_class_names}

\end{table}

\begin{table}[h] 
\centering 
\scalebox{1.0}{
\begin{tabular}{| l | c | c | }
 \hline \textbf{{Parent Class}}  &  \textbf{cv-index} & \textbf{Classes}  
 \\ \hline \hline grey8 & 0 & 5, m, 4, 8, 0, a, 1, k1
 \\ \hline grey8 & 1 & f, i, e, x , k8, k9 , k4, w
 \\ \hline grey8 & 2 & j, s , k5, d, k3, g , k, l
 \\ \hline grey8 & 3 & 2, p, u, v, 9, k2, n, y
 \\ \hline grey8 & 4 & q, r, 3, o , 7, z , k7, h
\\  \midrule
\end{tabular}}
\caption{ Partitioning of Data Sets for the grey8 classification problems. k\# denote the classes from the kmnist dataset.  } 
\label{tab:grey8_class_names}
\end{table}

\begin{table*}[h] 
\centering 
\scalebox{.70}{
\begin{tabular}{|l | l | l | }
 \hline \textbf{{Parent Class}}  &  \textbf{cv-index} & \textbf{Classes}  
\\ \hline \hline tiny2 & 0 & ox | slug
\\ \hline tiny2 & 1 & potpie | lawn mower, mower
\\ \hline tiny2 & 2 & sports car, sport car | lemon
\\ \hline tiny2 & 3 & dining table, board | volleyball
\\ \hline tiny2 & 4 & cannon | pop bottle, soda bottle
\\ \hline tiny2 & 5 & goldfish, Carassius auratus | magnetic compass
\\ \hline tiny2 & 6 & iPod | pole
\\ \hline tiny2 & 7 & space heater | computer keyboard, keypad
\\ \hline tiny2 & 8 & apron | wok
\\ \hline tiny2 & 9 & trilobite | maypole
\\ \hline tiny2 & 10 & suspension bridge | cash machine, cash dispenser, automated teller machine
\\ \hline tiny2 & 11 & vestment | rugby ball
\\ \hline tiny2 & 12 & sea cucumber, holothurian | golden retriever
\\ \hline tiny2 & 13 & black stork, Ciconia nigra | European fire salamander, Salamandra salamandra
\\ \hline tiny2 & 14 & pay-phone, pay-station | bullet train, bullet
\\ \hline tiny2 & 15 & go-kart | Labrador retriever
\\ \hline tiny2 & 16 & school bus | jellyfish
\\ \hline tiny2 & 17 & potter's wheel | Persian cat
\\ \hline tiny2 & 18 & orange | albatross, mollymawk
\\ \hline tiny2 & 19 & cliff dwelling | pill bottle
\\ \hline tiny2 & 20 & reel | hog, pig, grunter, squealer, Sus scrofa
\\ \hline tiny2 & 21 & guacamole | lion, king of beasts, Panthera leo
\\ \hline tiny2 & 22 & ice lolly, lolly, lollipop, popsicle | sulphur butterfly, sulfur butterfly
\\ \hline tiny2 & 23 & steel arch bridge | comic book
\\ \hline tiny2 & 24 & cougar, puma, mountain lion, painter, panther | picket fence, paling
\\ \hline tiny2 & 25 & broom | umbrella
\\ \hline tiny2 & 26 & lakeside, lakeshore | academic gown, academic robe, judge's robe
\\ \hline tiny2 & 27 & grasshopper, hopper | lifeboat
\\ \hline tiny2 & 28 & alp | moving van
\\ \hline tiny2 & 29 & wooden spoon | birdhouse
\\ \hline tiny2 & 30 & dumbbell | boa constrictor, Constrictor constrictor
\\ \hline tiny2 & 31 & mushroom | abacus
\\ \hline tiny2 & 32 & bison | stopwatch, stop watch
\\ \hline tiny2 & 33 & bucket, pail | espresso
\\ \hline tiny2 & 34 & neck brace | Arabian camel, dromedary, Camelus dromedarius
\\ \hline tiny2 & 35 & plunger, plumber's helper | lesser panda, red panda, panda
\\ \hline tiny2 & 36 & king penguin, Aptenodytes patagonica | gasmask, respirator, gas helmet
\\ \hline tiny2 & 37 & water tower | backpack, back pack, knapsack, packsack, rucksack, haversack
\\ \hline tiny2 & 38 & cliff, drop, drop-off | sunglasses, dark glasses, shades
\\ \hline tiny2 & 39 & pomegranate | bow tie, bow-tie, bowtie
\\ \hline tiny2 & 40 & desk | gondola
\\ \hline tiny2 & 41 & spiny lobster, langouste, rock lobster, crawfish, crayfish, sea crawfish | cockroach, roach
\\ \hline tiny2 & 42 & bannister, banister, balustrade, balusters, handrail | black widow, Latrodectus mactans
\\ \hline tiny2 & 43 & beer bottle | freight car
\\ \hline tiny2 & 44 & Chihuahua | frying pan, frypan, skillet
\\ \hline tiny2 & 45 & trolleybus, trolley coach, trackless trolley | scorpion
\\ \hline tiny2 & 46 & triumphal arch | torch
\\ \hline tiny2 & 47 & banana | altar
\\ \hline tiny2 & 48 & German shepherd, German shepherd dog, German police dog, alsatian | limousine, limo
\\ \hline tiny2 & 49 & crane | acorn
\\ \hline tiny2 & 50 & punching bag, punch bag, punching ball, punchball | bullfrog, Rana catesbeiana
\\ \hline tiny2 & 51 & snail | rocking chair, rocker
\\ \hline tiny2 & 52 & centipede | meat loaf, meatloaf
\\ \hline tiny2 & 53 & beaker | basketball
\\ \hline tiny2 & 54 & sock | butcher shop, meat market
\\ \hline tiny2 & 55 & bell pepper | ladybug, ladybeetle, lady beetle, ladybird, ladybird beetle
\\ \hline tiny2 & 56 & dugong, Dugong dugon | miniskirt, mini
\\ \hline tiny2 & 57 & fur coat | fountain
\\ \hline tiny2 & 58 & cauliflower | kimono
\\ \hline tiny2 & 59 & monarch, monarch butterfly, milkweed butterfly, Danaus plexippus | ice cream, icecream
\\ \hline tiny2 & 60 & swimming trunks, bathing trunks | water jug
\\ \hline tiny2 & 61 & chimpanzee, chimp, Pan troglodytes | sombrero
\\ \hline tiny2 & 62 & beach wagon, station wagon, wagon, estate car, waggon | dragonfly, darning needle
\\ \hline tiny2 & 63 & tailed frog, bell toad, ribbed toad, tailed toad, Ascaphus trui | thatch, thatched roof
\\ \hline tiny2 & 64 & tractor | beacon, lighthouse, beacon light, pharos
\\ \hline tiny2 & 65 & American lobster, Northern lobster, Maine lobster, Homarus americanus | turnstile
\\ \hline tiny2 & 66 & oboe, hautboy, hautbois | goose
\\ \hline tiny2 & 67 & CD player | guinea pig, Cavia cobaya
\\ \hline tiny2 & 68 & gazelle | flagpole, flagstaff
\\ \hline tiny2 & 69 & Egyptian cat | sewing machine
\\ \hline tiny2 & 70 & candle, taper, wax light | plate
\\ \hline tiny2 & 71 & teddy, teddy bear | pizza, pizza pie
\\ \hline tiny2 & 72 & projectile, missile | jinrikisha, ricksha, rickshaw
\\ \hline tiny2 & 73 & viaduct | pretzel
\\ \hline tiny2 & 74 & barbershop | orangutan, orang, orangutang, Pongo pygmaeus
\\ \hline tiny2 & 75 & confectionery, confectionary, candy store | coral reef
\\ \hline tiny2 & 76 & obelisk | binoculars, field glasses, opera glasses
\\ \hline tiny2 & 77 & convertible | spider web, spider's web
\\ \hline tiny2 & 78 & poncho | tarantula
\\ \hline tiny2 & 79 & brass, memorial tablet, plaque | barrel, cask
\\ \hline tiny2 & 80 & bathtub, bathing tub, bath, tub | hourglass
\\ \hline tiny2 & 81 & bee | chest
\\ \hline tiny2 & 82 & African elephant, Loxodonta africana | sea slug, nudibranch
\\ \hline tiny2 & 83 & police van, police wagon, paddy wagon, patrol wagon, wagon, black Maria | parking meter
\\ \hline tiny2 & 84 & cardigan | teapot
\\ \hline tiny2 & 85 & tabby, tabby cat | nail
\\ \hline tiny2 & 86 & bighorn, bighorn sheep, cimarron, Rocky Mountain bighorn, Rocky Mountain sheep, Ovis canadensis | snorkel
\\ \hline tiny2 & 87 & American alligator, Alligator mississipiensis | seashore, coast, seacoast, sea-coast
\\ \hline tiny2 & 88 & koala, koala bear, kangaroo bear, native bear, Phascolarctos cinereus | brain coral
\\ \hline tiny2 & 89 & sandal | organ, pipe organ
\\ \hline tiny2 & 90 & drumstick | bikini, two-piece
\\ \hline tiny2 & 91 & barn | chain
\\ \hline tiny2 & 92 & standard poodle | lampshade, lamp shade
\\ \hline tiny2 & 93 & walking stick, walkingstick, stick insect | remote control, remote
\\ \hline tiny2 & 94 & refrigerator, icebox | baboon
\\ \hline tiny2 & 95 & fly | scoreboard
\\ \hline tiny2 & 96 & Christmas stocking | syringe
\\ \hline tiny2 & 97 & Yorkshire terrier | dam, dike, dyke
\\ \hline tiny2 & 98 & mantis, mantid | brown bear, bruin, Ursus arctos
\\ \hline tiny2 & 99 & mashed potato | military uniform

\\  \midrule
\end{tabular}}
\caption{ Partitioning of the tiny-imagenet dataset. - separates the classes. } 

\label{tab:tiny2_class_names}
\end{table*}

\begin{table*}[h] 
\centering 
\scalebox{1.0}{
\begin{tabular}{l | c | c | c | c | c | c }
  &  \multicolumn{6}{c}{\textbf{Running-Times for A Single Forward Pass}} 
\\  \cmidrule(l){1-7} \textbf{Synthetic}  &  \multicolumn{2}{c}{\textbf{Local Features}} & \multicolumn{2}{c}{\textbf{Layer Features}} & \multicolumn{2}{c}{\textbf{Global Features }}  
\\ \textbf{2D-Data}   &  \multicolumn{2}{c}{ mean time (sec) } &  \multicolumn{2}{c}{ mean time (sec)}  & \multicolumn{2}{c}{ mean time (sec) } 
\\ \textbf{g-selected}   &  ph  & noph  &   ph  &  noph  &  ph    & noph 
\\ \textbf{Model}   &  $t^A$+$t^I$   &  $t^A$+$t^I$  &  $t^H$ &  $t^H$  &  $t^C$  &  $t^C$   
\\ \cmidrule(l){1-1} \cmidrule(l){2-7} 
synth fc6 & 1.575 $\pm$ 0.101 &  0.402 $\pm$ 0.022 &  0.026 $\pm$ 0.002 &  0.007 $\pm$ 0.001 &  0.443 $\pm$ 0.054 &  0.078 $\pm$ 0.009
\\ synth fc8 & 2.193 $\pm$ 0.192 &  0.561 $\pm$ 0.051 &  0.035 $\pm$ 0.012 &  0.009 $\pm$ 0.001 &  0.758 $\pm$ 0.095 &  0.114 $\pm$ 0.014
\\   synth fc10 & 2.901 $\pm$ 0.234 &  0.726 $\pm$ 0.043 &  0.044 $\pm$ 0.004 &  0.012 $\pm$ 0.001 &  1.195 $\pm$ 0.151 &  0.154 $\pm$ 0.019
\\ \cmidrule(l){1-1} \cmidrule(l){2-7} 
grey2 gconv2 & 2.248 $\pm$ 0.359 &  0.550 $\pm$ 0.270 &  0.056 $\pm$ 0.020 &  0.015 $\pm$ 0.020 &  1.384 $\pm$ 0.259 &  0.159 $\pm$ 0.022
\\ grey2 gconv3 & 1.302 $\pm$ 0.339 &  0.437 $\pm$ 0.301 &  0.053 $\pm$ 0.018 &  0.014 $\pm$ 0.012 &  0.413 $\pm$ 0.072 &  0.076 $\pm$ 0.015
\\  grey2 gconv4 & 2.615 $\pm$ 0.409 &  0.540 $\pm$ 0.319 &  0.059 $\pm$ 0.027 &  0.016 $\pm$ 0.014 &  0.252 $\pm$ 0.053 &  0.060 $\pm$ 0.015
\\ \cmidrule(l){1-1} \cmidrule(l){2-7} 
grey4 gconv2 & 2.308 $\pm$ 0.326 &  0.563 $\pm$ 0.276 &  0.057 $\pm$ 0.016 &  0.014 $\pm$ 0.014 &  1.526 $\pm$ 0.210 &  0.171 $\pm$ 0.023
\\ grey4 gconv3 & 1.385 $\pm$ 0.356 &  0.456 $\pm$ 0.317 &  0.056 $\pm$ 0.015 &  0.014 $\pm$ 0.013 &  0.506 $\pm$ 0.078 &  0.089 $\pm$ 0.020
\\   grey4 gconv4 & 2.711 $\pm$ 0.415 &  0.552 $\pm$ 0.330 &  0.060 $\pm$ 0.017 &  0.017 $\pm$ 0.014 &  0.335 $\pm$ 0.061 &  0.072 $\pm$ 0.021
\\ \cmidrule(l){1-1} \cmidrule(l){2-7} 
grey8 gconv2 & 2.348 $\pm$ 0.345 &  0.574 $\pm$ 0.284 &  0.059 $\pm$ 0.022 &  0.014 $\pm$ 0.014 &  1.714 $\pm$ 0.240 &  0.190 $\pm$ 0.035
\\ grey8 gconv3 & 1.388 $\pm$ 0.317 &  0.455 $\pm$ 0.309 &  0.056 $\pm$ 0.014 &  0.014 $\pm$ 0.013 &  0.629 $\pm$ 0.066 &  0.108 $\pm$ 0.027
\\   grey8 gconv4 & 2.779 $\pm$ 0.440 &  0.565 $\pm$ 0.339 &  0.064 $\pm$ 0.026 &  0.017 $\pm$ 0.014 &  0.482 $\pm$ 0.092 &  0.094 $\pm$ 0.027
\\ \cmidrule(l){1-1} \cmidrule(l){2-7} 
tiny2 tconv3 & 2.778 $\pm$ 0.609 &  0.786 $\pm$ 0.275 &  0.030 $\pm$ 0.005 &  0.008 $\pm$ 0.002 &  1.899 $\pm$ 0.711 &  0.191 $\pm$ 0.039
\\ tiny2 tconv4 &  2.423 $\pm$ 0.612 &  0.745 $\pm$ 0.281 &  0.032 $\pm$ 0.006 &  0.009 $\pm$ 0.002 &  1.202 $\pm$ 0.569 &  0.152 $\pm$ 0.036
\\   tiny2 tconv5 & 1.542 $\pm$ 0.314 &  0.597 $\pm$ 0.271 &  0.030 $\pm$ 0.003 &  0.010 $\pm$ 0.002 &  0.253 $\pm$ 0.072 &  0.059 $\pm$ 0.014
\\  \midrule
\end{tabular}}
\caption{Running times for different classes of topological features we have defined for a single forward pass.} 

\label{tab:running_times}

\end{table*}

\begin{table*}[p] 
\centering 
\scalebox{0.90}{
\begin{tabular}{l | c c c }
  &  \multicolumn{3}{c}{\textbf{Performance Estimation Basslines}} 
\\  \cmidrule(l){1-4} \textbf{Synthetic}  &  \multicolumn{1}{c}{\textbf{Model State}} & \multicolumn{1}{c}{\textbf{Testing Acc}} & \multicolumn{1}{c}{\textbf{Performance Gap}}
\\ \textbf{2D-Data}   &  \multicolumn{1}{c}{Mean Acc (\%)} & \multicolumn{1}{c}{ MAE (\%)} & \multicolumn{1}{c}{MAE (\%)}
\\ \cmidrule(l){1-1} \cmidrule(l){2-4} 
synth fc6 & 33.33  $\pm$  0.00 &  4.42 $\pm$ 4.83 &  4.19 $\pm$ 4.73  
\\ synth fc8 & 33.33  $\pm$  0.00  &  4.26 $\pm$ 4.50 &  4.05 $\pm$ 4.41
\\  synth fc10 & 33.33  $\pm$  0.00  &  4.09 $\pm$ 3.93  &  3.90 $\pm$ 3.82 
\\  \cmidrule(l){1-1} \cmidrule(l){2-4} \textbf{Grey 2}  &  \multicolumn{1}{c}{\textbf{Model State}} & \multicolumn{1}{c}{\textbf{Testing Acc}} & \multicolumn{1}{c}{\textbf{Performance Gap}}
\\ \textbf{Data}  &  \multicolumn{1}{c}{Mean Acc (\%)} & \multicolumn{1}{c}{ MAE (\%)} & \multicolumn{1}{c}{MAE (\%)}
\\ \cmidrule(l){1-1} \cmidrule(l){2-4} 
gconv2 & 33.33  $\pm$  0.00 &  4.89 $\pm$ 2.20  & 4.77 $\pm$ 2.32 
\\ gconv3 &  33.33  $\pm$  0.00 &  4.99 $\pm$ 2.45  &  5.09 $\pm$ 2.57 
\\ gconv4 &  33.33  $\pm$  0.00  &  5.41 $\pm$ 2.22  &  5.52 $\pm$ 2.35 
 \\  \cmidrule(l){1-1} \cmidrule(l){2-4} \textbf{Grey 4}  &  \multicolumn{1}{c}{\textbf{Model State}} & \multicolumn{1}{c}{\textbf{Testing Acc }} & \multicolumn{1}{c}{\textbf{Performance Gap}}
\\ \textbf{Data}  &  \multicolumn{1}{c}{Mean Acc (\%)} & \multicolumn{1}{c}{ MAE (\%)} & \multicolumn{1}{c}{MAE (\%)}
\\ \cmidrule(l){1-1} \cmidrule(l){2-4} 
gconv2 &  33.33  $\pm$  0.00 &  9.11 $\pm$ 2.45 &  8.65 $\pm$ 2.28 
\\ gconv3 &  33.33  $\pm$  0.00 &  9.51 $\pm$ 1.81 &  10.20 $\pm$ 2.34 
\\ gconv4 &  33.33  $\pm$  0.00  &  9.35 $\pm$ 1.73  &  9.83 $\pm$ 1.99 
 \\  \cmidrule(l){1-1} \cmidrule(l){2-4} \textbf{Grey 8}  &  \multicolumn{1}{c}{\textbf{Model State}} & \multicolumn{1}{c}{\textbf{Testing Acc }} & \multicolumn{1}{c}{\textbf{Performance Gap}}
\\ \textbf{Data}  &  \multicolumn{1}{c}{Mean Acc (\%)} & \multicolumn{1}{c}{ MAE (\%)} & \multicolumn{1}{c}{MAE (\%)}
\\ \cmidrule(l){1-1} \cmidrule(l){2-4}
gconv2 & 33.33  $\pm$  0.00  &  11.74 $\pm$ 1.99 &  12.64 $\pm$ 2.17 
\\ gconv3 &  33.33  $\pm$  0.00 &  12.61 $\pm$ 1.40  &  14.34 $\pm$ 1.69 
\\ gconv4 &  33.33  $\pm$  0.00 &  13.06 $\pm$ 1.30  &  14.29 $\pm$ 1.40 
 \\  \cmidrule(l){1-1} \cmidrule(l){2-4} \textbf{Tiny 2}  &  \multicolumn{1}{c}{\textbf{Model State}} & \multicolumn{1}{c}{\textbf{Testing Acc }} & \multicolumn{1}{c}{\textbf{Performance Gap}}
\\ \textbf{Data}  &  \multicolumn{1}{c}{Mean Acc (\%)} & \multicolumn{1}{c}{ MAE (\%)} & \multicolumn{1}{c}{MAE (\%)}
\\ \cmidrule(l){1-1} \cmidrule(l){2-4} 
tconv3 &  33.33  $\pm$  0.00 &  8.55 $\pm$ 4.49  &  11.42 $\pm$ 5.54 
\\ tconv4 &  33.33  $\pm$  0.00  &  8.37 $\pm$ 4.53  &  10.93 $\pm$ 5.84 
\\ tconv5 &  33.33  $\pm$  0.00 &  8.62 $\pm$ 4.92  &  11.55 $\pm$ 6.34   
\\  \midrule
\end{tabular}}
\caption{Performance Estimation Baselines. The first 1 column to the right of the model architecture names denotes the random chance classification accuracy to predict model states in predicting the intended model state using $\mathbb{NN}_k$. The next 2 columns present the mean absolute error in \% $\pm$ the standard error in estimating the testing accuracy and performance gap respectively using simply the median of the set of testing and performance gap values} 
\label{tab:performance_baselines_tab}
\label{tab:three}

\end{table*}

\begin{figure*}[bp!]
\begin{center}
 \includegraphics[width=1.0\linewidth]{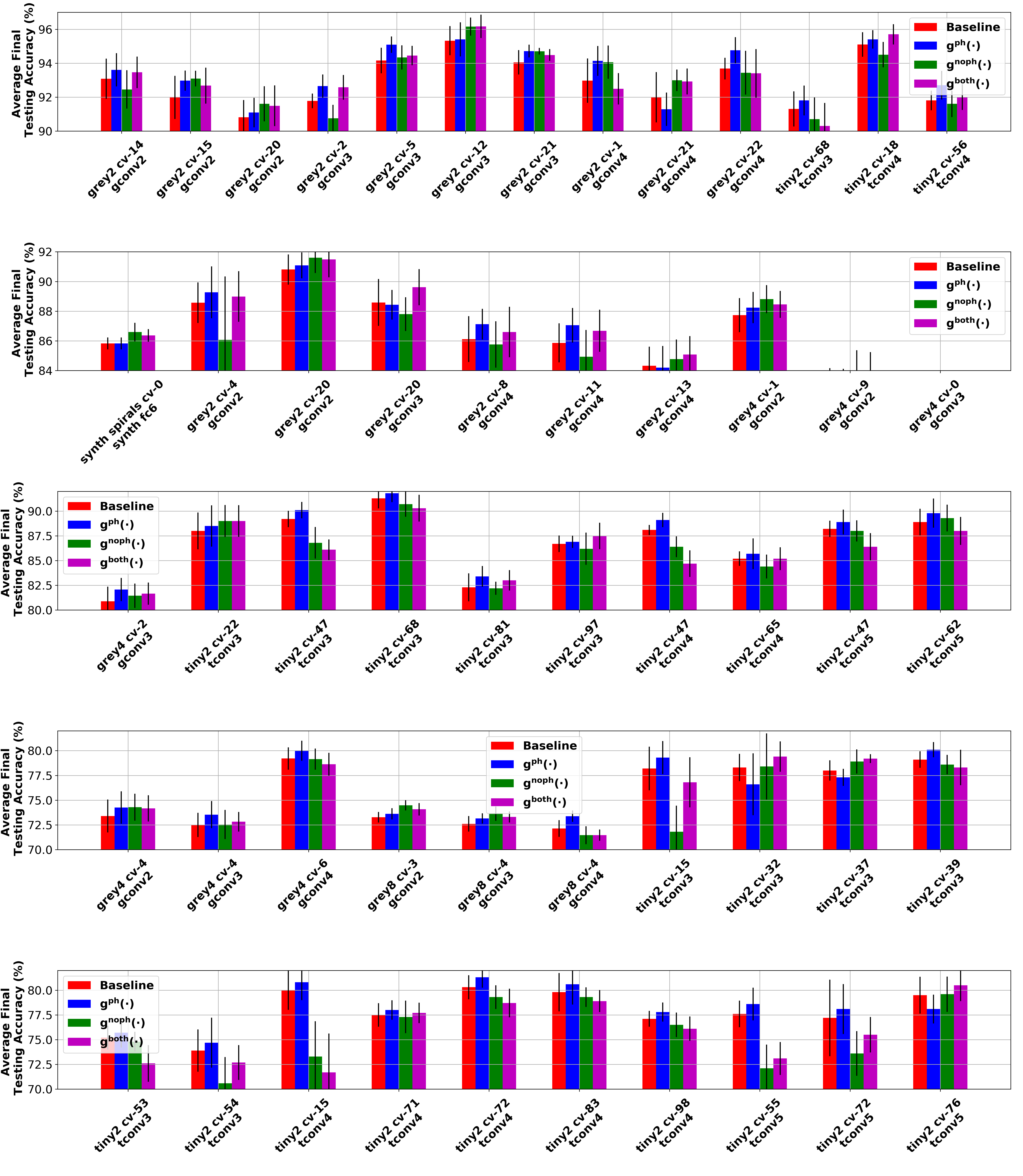}
\end{center}
  \caption{Additional results depict the effects of our meta-learning strategy on more challenging tasks. We similarly show there are classification problems where the topological meta-learning strategies consistently outperform the baseline training procedure denoted in red.}
\label{fig:meta_2}
\end{figure*}
